\documentclass{article}

\PassOptionsToPackage{numbers, compress}{natbib}

\usepackage[final]{neurips_data_2021}

\usepackage[utf8]{inputenc} 
\usepackage[T1]{fontenc}    
\usepackage{hyperref}       
\usepackage{url}            
\usepackage{booktabs}       
\usepackage{amsfonts}       
\usepackage{nicefrac}       
\usepackage{microtype}      

\usepackage{graphicx}
\usepackage{caption}
\usepackage[dvinames,dvipsnames]{xcolor}
\usepackage{multirow}
\usepackage{amsmath,amssymb,amsfonts,dsfont}  
\usepackage{bm} 
\usepackage{comment}
\usepackage{enumitem}
\usepackage{pifont}
\usepackage{tikz}
\usepackage{booktabs}
\usepackage{nicefrac}

\newtheorem{definition}{Definition}

\DeclareMathOperator*{\argmin}{arg\,min}

\newcommand{\note}[1]{
	\noindent~\\
	\fcolorbox{red}{orange}{\parbox{0.99\columnwidth}{#1}}
}
\renewcommand{\note}[1]{}

\newcommand{\nopts}{13}
\newcommand{\noptsfams}{6}
\newcommand{\nbenchs}{22}
\newcommand{\nbenchfams}{12}
\newcommand{\ncommunitybenchfams}{7}
\newcommand{\nnewbenchfams}{5}
\newcommand{\bb}{black-box}
\newcommand{\mf}{multi-fidelity}

\newcommand{\rescaptionexisting}[1]{#1 (lower is better). We report median performance (regret for tabular/surrogate benchmarks and function values for raw benchmarks) across $32$ repetitions per \emph{existing community} benchmark.}

\newcommand{\rescaptionnew}[2]{#1 (lower is better). We report the median normalized regret across $32$ repetitions for each \emph{new} benchmarks collected for #2.}

\newcommand{\stattestcaption}{ We boldface the best result per row.}

\newcommand{\wrt}{w.r.t.~}

\newcommand{\dimpcs}[0]{d}
\newcommand{\pcs}[0]{\bm{\Lambda}}

\newcommand{\conf}[0]{\bm{\lambda}}

\newcommand{\bo}{\textit{BO}}
\newcommand{\smac}{\textit{SMAC}}
\newcommand{\smachb}{\textit{SMAC-HB}}
\newcommand{\smacrf}{\textit{BO$_{RF}$}}
\newcommand{\smacbo}{\textit{BO$_{GP}$}}
\newcommand{\hb}{\textit{HB}}
\newcommand{\sh}{\textit{SH}}
\newcommand{\tpe}{\textit{BO$_{KDE}$}}
\newcommand{\bohb}{\textit{BOHB}}
\newcommand{\dragonfly}{\textit{DF}}
\newcommand{\dehb}{\textit{DEHB}}
\newcommand{\de}{\textit{DE}}
\newcommand{\random}{\textit{RS}}
\newcommand{\rayha}{\textit{Ray$_{hyp}^{asha}$}}
\newcommand{\raytpe}{\textit{Ray$_{hyp}$}}
\newcommand{\optunatm}{\textit{Optuna$_{tpe}^{md}$}}
\newcommand{\optunath}{\textit{Optuna$_{tpe}^{hb}$}}
\newcommand{\hebo}{\textit{HEBO}}

\newcommand{\pybnn}{\textit{BNN}}
\newcommand{\paramnettime}{\textit{Net}}
\newcommand{\svm}{\textit{SVM}}
\newcommand{\lr}{\textit{LogReg}}
\newcommand{\rf}{\textit{RandomForest}}
\newcommand{\xgb}{\textit{XGBoost}}
\newcommand{\mlp}{\textit{MLP}}

\newcommand{\NASHPO}{\textit{NBHPO}}
\newcommand{\NASOOO}{\textit{NB101}}
\newcommand{\NASOSO}{\textit{NB1Shot1}}
\newcommand{\NASTOO}{\textit{NB201}}

\newcommand{\cartpole}{\textit{Cartpole}}

\newcommand{\NASOSOA}{\textit{NB1Shot1}$_{1}$}
\newcommand{\NASOSOB}{\textit{NB1Shot1}$_{2}$}
\newcommand{\NASOSOC}{\textit{NB1Shot1}$_{3}$}

\newcommand{\NASA}{\textit{NB101}$_{Cf10A}$}
\newcommand{\NASB}{\textit{NB101}$_{Cf10B}$}
\newcommand{\NASC}{\textit{NB101}$_{Cf10C}$}
\newcommand{\slice}{\textit{NBHPO}$_{Slice}$}
\newcommand{\protein}{\textit{NBHPO}$_{Prot}$}
\newcommand{\naval}{\textit{NBHPO}$_{Naval}$}
\newcommand{\parkinson}{\textit{NBHPO}$_{Park}$}

\newcommand{\nbcifartv}{\textit{NB201}$_{Cf10V}$}
\newcommand{\nbcifarh}{\textit{NB201}$_{Cf100}$}
\newcommand{\nbimage}{\textit{NB201}$_{INet}$}
\newcommand{\bnnprotein}{\textit{BNN}$_{Protein}$}
\newcommand{\bnnyear}{\textit{BNN}$_{Year}$}
\newcommand{\paramadult}{\textit{Net}$_{Adult}$}
\newcommand{\paramhiggs}{\textit{Net}$_{Higgs}$}
\newcommand{\paramletter}{\textit{Net}$_{Letter}$}
\newcommand{\parammnist}{\textit{Net}$_{MNIST}$}
\newcommand{\paramoptdigits}{\textit{Net}$_{OptDig}$}
\newcommand{\parampoker}{\textit{Net}$_{Poker}$}

\newcommand{\hpobench}{\textit{HPOBench}}
\newcommand{\automl}{\textit{AutoML}}

\newcommand{\cmark}{\ding{51}}%
\newcommand{\xmark}{\ding{55}}%
\newcommand{\yesmark}{{\color{ForestGreen}\cmark}}
\newcommand{\nomark}{{\color{BrickRed}\xmark}}
\newcommand{\maybemark}{{\color{NavyBlue}(\cmark)}}

\title{HPOBench: A Collection of Reproducible Multi-Fidelity Benchmark Problems for HPO}

\author{%
  Katharina Eggensperger$^1$\thanks{\{eggenspk,mallik,fh\}@cs.uni-freiburg.de}~, Philipp Müller$^1$, Neeratyoy Mallik$^1$, Matthias Feurer$^1$, \\
  \textbf{René Sass$^2$, Aaron Klein$^3\thanks{work done prior to joining Amazon}$~, Noor Awad$^1$, Marius Lindauer$^2$, Frank Hutter$^{1,4}$} \\
  $^1$ Albert-Ludwigs-Universität Freiburg
  $^2$ Leibniz Universität Hannover \\
  $^3$ Amazon
  $^4$ Bosch Center for Artificial Intelligence
}

\begin{document}

\maketitle

\begin{abstract}
  To achieve peak predictive performance, hyperparameter optimization (HPO) is a crucial component of machine learning and its applications. Over the last years, the number of efficient algorithms and tools for HPO grew substantially. At the same time, the community is still lacking 
  realistic, diverse, computationally cheap, and standardized benchmarks. This is especially the case for multi-fidelity HPO methods.
  To close this gap, we propose \hpobench{}, which includes $\ncommunitybenchfams$ existing and $\nnewbenchfams$ new benchmark families, with a total of more than $100$ \mf{} benchmark problems. \hpobench{} allows to run this extendable set of \mf{} HPO benchmarks in a reproducible way by isolating and packaging the individual benchmarks in containers. It also provides surrogate and tabular benchmarks for computationally affordable yet statistically sound evaluations. To demonstrate \hpobench{}'s broad compatibility with various optimization tools, as well as its usefulness, we conduct an exemplary large-scale study evaluating 
  $\nopts$ optimizers from $\noptsfams$ optimization tools.
We provide \hpobench{} here: \url{https://github.com/automl/HPOBench}.
\end{abstract}

\section{Introduction}
\label{sec:intro}

The plethora of design choices in modern machine learning (ML) makes research on practical and effective methods for hyperparameter optimization (HPO) ever more important. In particular, ever-growing
models and datasets create a demand for new HPO methods that are more efficient and powerful than existing \bb{} optimization (BBO) methods. Especially if it is only feasible to evaluate very few models fully, \mf{} optimization methods have been shown to yield impressive results by trading off cheap-to-evaluate proxies and expensive evaluations on the real target~\citep{forrester-prs07a,klein-aistats17,kandasamy-icml17a,jamieson-aistats16a,awad-ijcai21}. They showed tremendous speedups, such as accelerating the search process in low-dimensional ML hyperparameter spaces by a factor of $10$ to $1000$~\cite{klein-aistats17,awad-ijcai21}. However, the development of such methods often happens in isolation, which potentially prevents HPO research from reaching its full potential.
Prior publications on new HPO methods (i)~often relied on artificial test functions and low-dimensional toy problems, (ii)~sometimes introduced a new set of problems, (iii)~set up on different computing environments, having different requirements and interfaces, and (iv)~often did not open-source their code base.
All of these make it difficult to compare and develop methods, necessitating an evolving set of relevant and up-to-date benchmark problems which drives continued and quantifiable progress in the community.

While there are efforts to simplify benchmarking HPO and global optimization algorithms~\citep{eggensperger-bayesopt13,doerr-arxiv2018a,bayesmark,hansen-oms20,nevergrad,bliek-arxiv21a,haese-MLST21}, we are not aware of efforts to collect a diverse set of benchmarks in a single library, with a unified interface and countering potentially conflicting dependencies that may arise over time. The latter is particularly important because the rapid evolution of the Python-ML ecosystem can render a benchmark no longer usable for the community after a major release was published. This creates a significant hurdle for contribution from the community to grow a benchmark library.
To solve this issue, we propose~\hpobench{}, a benchmark suite for HPO problems, with a special focus on \mf{} problems, licensed under a permissive OSS license (\emph{Apache 2.0}) and available at \url{https://github.com/automl/HPOBench}. \hpobench{} provides a common interface and an infrastructure to isolate benchmarks in their own containers and implements $\nbenchfams$ popular benchmark families, each with multiple problems and preserved with its dependencies in a container for long-term use. To enable efficient comparisons, most of these benchmarks are table- or surrogate-based, enabling resource efficient large-scale experiments, which we demonstrate in this work. Our contributions are:
\begin{enumerate}
    \item The first available collection of \mf{} HPO problems. It contains $\nbenchfams$ benchmark families with $100$+ \mf{} HPO problems under a unified interface, comprising traditional HPO and neural architecture search (NAS). These benchmarks also define the largest collection of \bb{} HPO problems to date.
    \item The first collection of \textit{containerized} benchmarks to ensure the longevity, maintainability and extensibility of benchmarks.
    \item The first set of HPO benchmarks that are available as both, the \emph{raw} benchmark and the \emph{tabular} version.
    \item The first HPO benchmark that also supports multi-objective optimization and transfer-HPO across datasets (and arbitrary combinations of these with multiple fidelities). 
    \item We demonstrate how \hpobench{} can be used in an exemplary large-scale study with $\nopts$ optimizers from $\noptsfams$ optimization tools, assessing whether advanced methods outperform random search and how effective multi-fidelity HPO is.
\end{enumerate}

This paper is structured as follows. We first discuss background on HPO and \mf{} optimization (Section \ref{sec:background}). Then, we discuss related work on benchmarking 
(Section \ref{sec:related}). Next, we describe the challenges for an HPO benchmark and how \hpobench{} alleviates them (Section~\ref{sec:hpobench}). Then, we conduct a large-scale comparison of existing, popular HPO methods to demonstrate the usefulness of \hpobench{} (in Section~\ref{sec:exp}). We conclude the paper by highlighting further advantages and potential future work (Section~\ref{sec:discussion}).

\section{Background on Hyperparameter Optimization}
\label{sec:background}

With \hpobench{} we aim to provide benchmarks to evaluate HPO methods. In the following, we briefly formalize BBO for HPO and survey \mf{} optimization (see Feurer and Hutter~\citep{feurer-bookchapter19a} for a detailed overview), both with a focus on the methods used in our experiments.

\subsection{Black-box Hyperparameter Optimization}
\label{ssec:back:hpo}

Black-box optimization (BBO) aims to find a solution $\argmin_{\conf \in \pcs} f(\conf)$ where $f$ is a \bb{} function, for which typically no gradients are available, we cannot make any statements about its smoothness, convexity and noise level. In summary, the only mode of interaction with \bb{} functions is querying them at given inputs $\conf$ and measuring the quantity of interest $f(\conf)$. In the context of HPO, $\conf \in \pcs$ is a hyperparameter configuration where the domain $\Lambda_i$ of a hyperparameter is often bounded and continuous, but can also be integer, ordinal or categorical. There are also so-called conditional hyperparameters~\citep{bergstra-nips11a,thornton-kdd13a} defining hierarchical search spaces; however, the first version of \hpobench{} focuses on flat configuration spaces first as all optimizers support this.

There are three broad families of BBO methods: (i) purely explorative approaches such as Random Search (\random{}) and grid search 
are simple but sample-inefficient; (ii) model-free Evolutionary Algorithms (EAs) 
based on mutation, crossover and selection operators applied to a population of configurations require comparably large resources to evaluate the entire population but can perform very well given enough resources;
(iii) iterative model-based methods, such as Bayesian Optimization~\citep{shahriari-ieee16a}, which are guided by a predictive model trained on prior function evaluations are known as the most sample-efficient methods. We include representative algorithms from each of these 3 families in our exemplary experiments in Section \ref{sec:exp}.

\subsection{Multi-fidelity Hyperparameter Optimization}
\label{ssec:back:mf}

To efficiently optimize today's ever-growing ML models, multi-fidelity approaches
relax the \bb{} assumption by allowing cheaper queries at lower fidelities $b$ as well ($\argmin_{\conf \in \pcs} f(\conf, b)$). Examples for these approximations include dataset subsets~\citep{klein-aistats17,petrak-tr00a,swersky-nips13a}, feature subsets~\citep{li-jmlr18a} or lower number of epochs~\citep{li-jmlr18a,swersky-arxiv14a,domhan-ijcai15a}. Multi-fidelity methods have been shown to lead to speedups of up to $1000\times$ over \bb{} methods~\cite{klein-aistats17,awad-ijcai21}.
\hpobench{} will allow the community to compare different \mf{} methods and in the following we give an overview of representative methods. 

A popular \mf{} HPO approach that discretizes the fidelity space is Hyperband (\hb{}~\citep{li-jmlr18a}), a very simple method with strong empirical performance. It randomly samples new configurations and allocates more resources to promising configurations by repeatedly calling successive halving (\sh{}~\cite{jamieson-aistats16a}) as a sub-algorithm. 
The simplicity and effectiveness of \hb{} have been leveraged with other popular \bb{} optimizers for improved performance:
\bohb{}~\cite{falkner-icml18a} combines \hb{} with Bayesian Optimization (\bo{}) and DEHB~\cite{awad-ijcai21} combines it with the evolutionary approach of Differential Evolution (\de{}~\citep{storn-jgo1997a,awad-iclr20}).
The non-\hb{}-based multi-fidelity case has also been researched extensively~\citep{klein-aistats17,kandasamy-icml17a,swersky-nips13a,swersky-arxiv14a,domhan-ijcai15a,golovin-kdd17a,wu-uai20a,takeno-icml20a,song-aistats19a}. Not being limited to predefined fidelity values makes these methods very powerful, but they rely on strong models to avoid poor choices of fidelities, often making HB-based fidelity selection more robust. 
To study the efficacy of \mf{} optimization, in our exemplary experiments in Section \ref{sec:exp}, we primarily compared \bb{} optimizers against their \mf{} versions (i.e., \random{} vs. \hb{}, \bo{} vs. \bohb{}, and \de{} vs. \dehb{}). These experiments show large speedups of multi-fidelity optimizers in the regime of small compute budgets, whereas for large compute budgets multi-fidelity optimization is less useful.

Besides multi-fidelity optimization, a very active field of study to speed up HPO is to use transfer-learning across datasets~\citep{vanschoren-automl19a,bardenet-icml13a,yogatama-aistats14a,feurer-aaai15a,wistuba-ml18a}; we note that transfer HPO methods can also be evaluated with \hpobench{} by learning across the datasets within each of its families.

\section{Related Work}
\label{sec:related}

Proper benchmarking is hard. It is important to be aware of technical and methodological pitfalls, e.g. comparing implementations instead of algorithms~\citep{kriegel-inf2017a,narang-arxiv2021a}, comparing tuned algorithms versus untuned baselines~\citep{bergstra-icml13a, melis-iclr18a}, to not fall for an illusion of progress~\citep{hand-ss2006a,dacrema-recsys2019a} and to know which sources of variance exist and control for them~\citep{bouthillier-mlsys21a}. Also, there is a rich literature on how to empirically evaluate and compare methods in various domains, e.g. evolutionary optimization~\citep{weise-jcst12a}, planning~\citep{howe-jair02a}, satisfiability and constraint satisfaction~\citep{gent-report97a}, algorithm configuration~\citep{eggensperger-jair19a}, NAS~\citep{lindauer-jmlr20a}, and also for benchmarking optimization algorithms~\citep{bartz-beielstein-arxiv2020v2}. Our goal is not to provide further recommendations on how and why to benchmark, but to provide concrete benchmarks to simplify development and to improve the reproducibility and comparability of HPO and in particular \mf{} methods.

Furthermore, there have been a lot of efforts to provide optimization benchmarks for the community. Having a common set of benchmark problems in a unified format fosters and guides research. Prominent examples in the area of HPO are \href{https://bitbucket.org/mlindauer/aclib2/src/master/}{ACLib}~\citep{hutter-lion14a} for algorithm configuration, \href{https://github.com/numbbo/coco}{COCO}~\citep{hansen-oms20} for continuous optimization,  \href{https://github.com/uber/bayesmark}{Bayesmark}~\citep{bayesmark} for Bayesian optimization, \href{https://github.com/aspuru-guzik-group/olympus}{Olympus}~\citep{haese-MLST21} for optimization of experiment planning tasks, and \href{https://github.com/releaunifreiburg/HPO-B}{HPO-B}~\citep{pineda-neurips21a} for transfer-HPO methods (for more, see Appendix~\ref{app:related:otherframeworks}). However, no benchmark so far has \mf{} optimization problems, supports preserving a diverse set of benchmarks for the longer term (containers), supports multiple objectives, and provides cheap-to-evaluate surrogate/tabular benchmarks; we hope to close this gap with \hpobench{}.

Besides benchmarks, competitions are another form of focusing research effort by providing a common goal and incentive. Famous examples are the \automl{} challenges~\cite{guyon-automl19a}, the \emph{AutoDL} challenge~\cite{liu-hal20a}, the \href{https://numbbo.github.io/workshops/index.html}{GECCO BBOB workshop series} based on COCO~\citep{hansen-oms20} and the NeurIPS 2020 BBO challenge~\citep{turner-neuripscomp20a} (for more, see Appendix~\ref{app:related:competitions}). In contrast to these, we do not focus on defining concrete experimentation protocols, but rather on providing a flexible benchmarking environment to study, develop and compare optimization methods.

\section{HPOBench: A Benchmark Suite for Multi-Fidelity Hyperparameter Optimization benchmarks}
\label{sec:hpobench}

In this section, we present \hpobench{}, a collection of HPO benchmarks defined as follows:

\begin{definition}[HPO Benchmark]\label{def:bench} An HPO benchmark consists of a function $f : \conf \xrightarrow{} \mathcal{R}$ to be minimized and a (bounded) hyperparameter space $\pcs$ with hyperparameters $[\Lambda_1, \dots, \Lambda_{\dimpcs}]$ of type continuous, integer, categorical or ordinal. In the case of \mf{} benchmarks, $f$ can be queried at lower fidelities, $f : \conf \times \bm{b} \xrightarrow{} \mathcal{R}$, and the fidelity space $\bm{B}$ describes which low-fidelities $[B_{1}, \dots, B_{e}]$ of type continuous, integer or ordinal are available.
\end{definition}

Specifically, each benchmark consists of the implementation of that function, which returns at least one loss. Since this function typically evaluates an ML algorithm, the benchmark defines all relevant settings, dependencies and inputs, such as datasets, splits and how to compute the loss.

In the remainder of this section, we first discuss the desiderata of a benchmark that aids HPO research and then highlight the features of \hpobench{} by detailing how its design fulfills these desiderata.

\subsection{Desiderata of an HPO Benchmark}
\label{ssec:hpo:comm}

One of the challenges posed to standardized HPO research lies in the varied choices of the underlying ML components -- datasets and their splits, preprocessing, hyperparameter ranges, underlying software versions, and hardware used. Moreover, the practices applied in HPO research itself can vary along the lines of optimization budget, number of repetitions, metrics measured and reported. This leads to inconsistencies and difficulties in comparison of different HPO methods across publications and over time, affecting the reproducibility of experiments that hinders continued progress in HPO research. 

In order to alleviate such issues and encourage participation by the research community, benchmarks need to standardize these practices to allow the community to be an active stakeholder in developing and re-using benchmarks. \hpobench{} is designed to both allow easy, flexible use with a minimal API that is identical for all benchmarks (see Figure~\ref{fig:code}); and have a low barrier for contributing new benchmark problems. We, therefore, identify $3$ features of a benchmark that allow its wide-scale use and long-term applicability: (i) \textit{efficiency} by providing tabular and surrogate benchmarks for quick, efficient experiments, along with the original benchmarks; (ii) \textit{reproducibility} of results by containerizing benchmarks; and (iii) \textit{flexibility} by covering different optimization landscapes and possible use cases, e.g. multi-objective, transfer-HPO, and even multi-fidelity optimization with multiple fidelity variables. To our knowledge, no other existing benchmarks offer these possibilities.
\hpobench{} provides a framework to enable standardized, principled research and experimentation. 
We list all benchmarks that are included in \hpobench{} in Table~\ref{tab:benchmarks} and provide a detailed description of the respective configuration spaces in Appendix~\ref{app:benchmarks}.

\subsection{Efficiency}
\label{ssec:hpobench:efficiency}

HPO benchmarks that follow Definition~\ref{def:bench} exhibit the drawback that they evaluate a costly function, rendering the empirical comparison of optimization algorithms expensive and ruling out such benchmarks for interactive development of new methods. To overcome this issue, beside such raw benchmarks, we also provide two well-established benchmark classes which alleviate this issue:

\begin{definition}[Tabular Benchmark]\label{def:tab_bench}
A tabular benchmark returns values from a lookup table with recorded function values of a raw HPO benchmark instead of evaluating $f(\conf)$. The (bounded) hyperparameter space is restricted to only contain these values and therefore bears a form of discretization. 
In the case of \mf{} benchmarks, each tabular benchmark has a fidelity space and the underlying table also contains the recorded function values on the low-fidelities.
\end{definition}

Tabular benchmarks are popular in the HPO community as they are easy to distribute and induce little overhead~\citep{snoek-nips12a,bardenet-icml13a,wistuba-ecml16a,ying-icml19a,metz-arxiv2020a}, however, they require to discretize the hyperparameter space.
Surrogate benchmarks~\citep{eggensperger-aaai15,eggensperger-mlj18a} are an alternative since they provide the original hyperparameter space.

\begin{definition}[Surrogate Benchmark]\label{def:surr_bench}
A surrogate benchmark returns function values predicted by an ML model trained on a tabular benchmark or recorded function values of a raw HPO benchmark. It reuses the original hyperparameter space and can be extended to the \mf{} case as well.
\end{definition}

While surrogate benchmarks are similarly cheap to query, the surrogate's internal ML model adds extra complexity and the benchmark's quality crucially depends on the quality of this model and its training data. Because surrogate benchmarks yield a drop-in replacement for raw benchmarks, they enjoy widespread adoption in the HPO community~\citep{falkner-icml18a,perrone-neurips18a,eggensperger-mlj18a,martinez-cantin-ieee2019a,klein-neurips19a,daxberger-ijcai20a,siems-arxiv20a}.

Furthermore, while \hpobench{} puts a strong focus on \mf{} benchmarks, it also facilitates evaluating \bb{} optimization algorithms. In fact, a \mf{} benchmark with $k$ different fidelity levels can be used to define $k$ separate (yet related) benchmarks for \bb{} optimization. As such, \hpobench{} defines more than $400$ \bb{} HPO benchmarks.

\subsection{Reproducibility}
\label{ssec:hpobench:reproducibility}

One of the challenges that come with many new benchmarks is their one-off development and their lack of maintenance. This means that any new update to the benchmark or its dependencies can easily lead to conflicts and inconsistencies with respect to software dependencies and possibly old published results (see Appendix~\ref{app:dependencies} for examples). While in practice the very same problem, also known as \emph{dependency hell}, can also occur on the optimizer side, in this paper we focus on the benchmark side. 

\begin{figure}
 \centering
\begin{minipage}[b]{0.6\textwidth}
    \!\!\!\!\!\!\!\!\!\!\!\!\!\!\!\resizebox{1.15\textwidth}{!}{%
    \usetikzlibrary{calc, fit, positioning}
\definecolor{green}{RGB}{27,158,119}
\definecolor{orange}{RGB}{217,95,2}
\definecolor{purple}{RGB}{117,112,179}
\definecolor{pink}{RGB}{231,41,138}
\begin{tikzpicture}[%
    node distance=10em,
    box/.style={rectangle,
                draw=black,
                thick,
                fill=white,
                text width=7em,
                align=center,
                rounded corners,
                text centered,
                minimum height=1em,
    },
    box_benchmark/.style={box, draw=white,
                          fill=orange!50
    },
    box_data/.style={box, draw=white,
                     fill=green!50
    },
    box_requirements/.style={box, draw=white,
                             fill=pink!50
    },
    dotted_box/.style={rectangle,
                      draw=black,
                      thick,
                      fill=gray!20,
                      text width=5em,
                      align=center,
                      text centered,
                      minimum height=1em
    },
]
\node (dataA) [box_benchmark] {Data set $\alpha$};
\node (dataB) [box_benchmark, below of = dataA, node distance=5em] {Data set $\beta$};

\node (reqA) [box_requirements, right of = dataA] {Requirements a};
\node (reqB) [box_requirements, right of = dataB] {Requirements b};

\node (benchA) [box_data, right of = reqA] {Benchmark A};
\node (benchB) [box_data, right of = reqB] {Benchmark B};

\node (pythonAPI_1) [box, right of = benchA, xshift=2em, yshift=-2.5em, minimum height=4em,text width=4em, fill=purple!50, draw=white] {Python\\API};

\node (OPT) [box, right of = pythonAPI_1, minimum height=4em, text width=4em, fill=purple!50, draw=white, node distance=3cm] {Optimizer};

\draw [<->, thick] (OPT.west) to (pythonAPI_1.east);
\draw [-, dotted, thick] (dataA.east) to (reqA.west);
\draw [-, dotted, thick] (reqA.east) to (benchA.west);
\draw [-, dotted, thick] (dataB.east) to (reqB.west);
\draw [-, dotted, thick] (reqB.east) to (benchB.west);

\draw[line width=0.05cm, rounded corners, draw=black]     ($(dataA.north west)+(-0.5,0.5)$) rectangle ($(benchA.south east)+(0.5,-0.25)$);
\draw[line width=0.05cm, rounded corners, draw=black]     ($(dataB.north west)+(-0.5,0.5)$) rectangle ($(benchB.south east)+(0.5,-0.25)$);

\draw[line width=0.05cm, rounded corners, draw=black]     ($(pythonAPI_1.north west)+(-0.25,0.5)$) rectangle ($(OPT.south east)+(0.25,-0.25)$);

\node (containerA) [dotted_box, above of = benchA, text width=3cm, yshift=-8.em, xshift=1em] {Container Env};
\node (containerB) [dotted_box, above of = benchB, text width=3cm, yshift=-8.em, xshift=1em] {Container Env};
\node (pythonEnv)  [dotted_box, above of = OPT, node distance=3.2em, text width = 7em, xshift=2.em] {Python Env};

\draw[->, thick]      ($(benchA.east)+(0.5, 0.)$) -- ($(pythonAPI_1.west)+(0.0, 0.1)$);
\draw[->, thick]      ($(benchB.east)+(0.5, 0)$)  --  ($(pythonAPI_1.west)+(0.0, -0.15)$);
\draw[<-, thick]      ($(benchA.east)$)  --  ($(benchA.east)+(0.5, 0.0)$);
\draw[<-, thick]      ($(benchB.east)$)  --  ($(benchB.east)+(0.5, 0.0)$);

\draw[very thick, dashed]     ($(dataB.west)+(-1., -0.9)$)    -- ($(benchB.east)+(8.0, -0.9)$);

\node (data2_A) [box_benchmark, below of = dataA, yshift=-1.1em] {Data set $\alpha$};
\node (data2_B) [box_benchmark, below of = data2_A, yshift=8em] {Data set $\beta$};

\node (req2) [box_requirements, right of = data2_A, yshift=-1em] {Requirements a $\bigcup$ b};

\node (bench2_A) [box_data, right of = req2, yshift=+1em] {Benchmark A};
\node (bench2_B) [box_data, below of = bench2_A, yshift=8em] {Benchmark B};

\node (pythonAPI_2) [box, right of = bench2_A, xshift=2em, yshift=-1em, minimum height=4em,text width=4em, fill=purple!50, draw=white] {Python\\API};

\node (OPT2) [box, right of = pythonAPI_2, minimum height=4em, text width=4em, fill=purple!50, draw=white, node distance=3cm] {Optimizer};

\draw [<->, thick] (OPT2.west) to (pythonAPI_2.east);
\draw [<->, thick] (pythonAPI_2.west) to (bench2_A.east);
\draw [<->, thick] (pythonAPI_2.west) to (bench2_B.east);

\draw [-, dotted, thick] (bench2_A.west) to (req2.east);
\draw [-, dotted, thick] (bench2_B.west) to (req2.east);
\draw [-, dotted, thick] (data2_A.east) to (req2.west);
\draw [-, dotted, thick] (data2_B.east) to (req2.west);

\draw[line width=0.05cm, rounded corners, draw=black]     ($(data2_A.north west)+(-0.5,0.5)$) rectangle ($(OPT2.south east)+(0.5,-0.3)$);

\node (pythonEnv_2)  [dotted_box, above of = OPT2, text width=7em, node distance=3.2em, xshift=2.em] {Python Env};

\end{tikzpicture}
    }%
    \caption{Overview of benchmark environments with (upper) and without (lower) using containers.}
    \label{fig:container}
\end{minipage}
\qquad
\begin{minipage}[b]{0.33\textwidth}
\includegraphics[width=\textwidth]{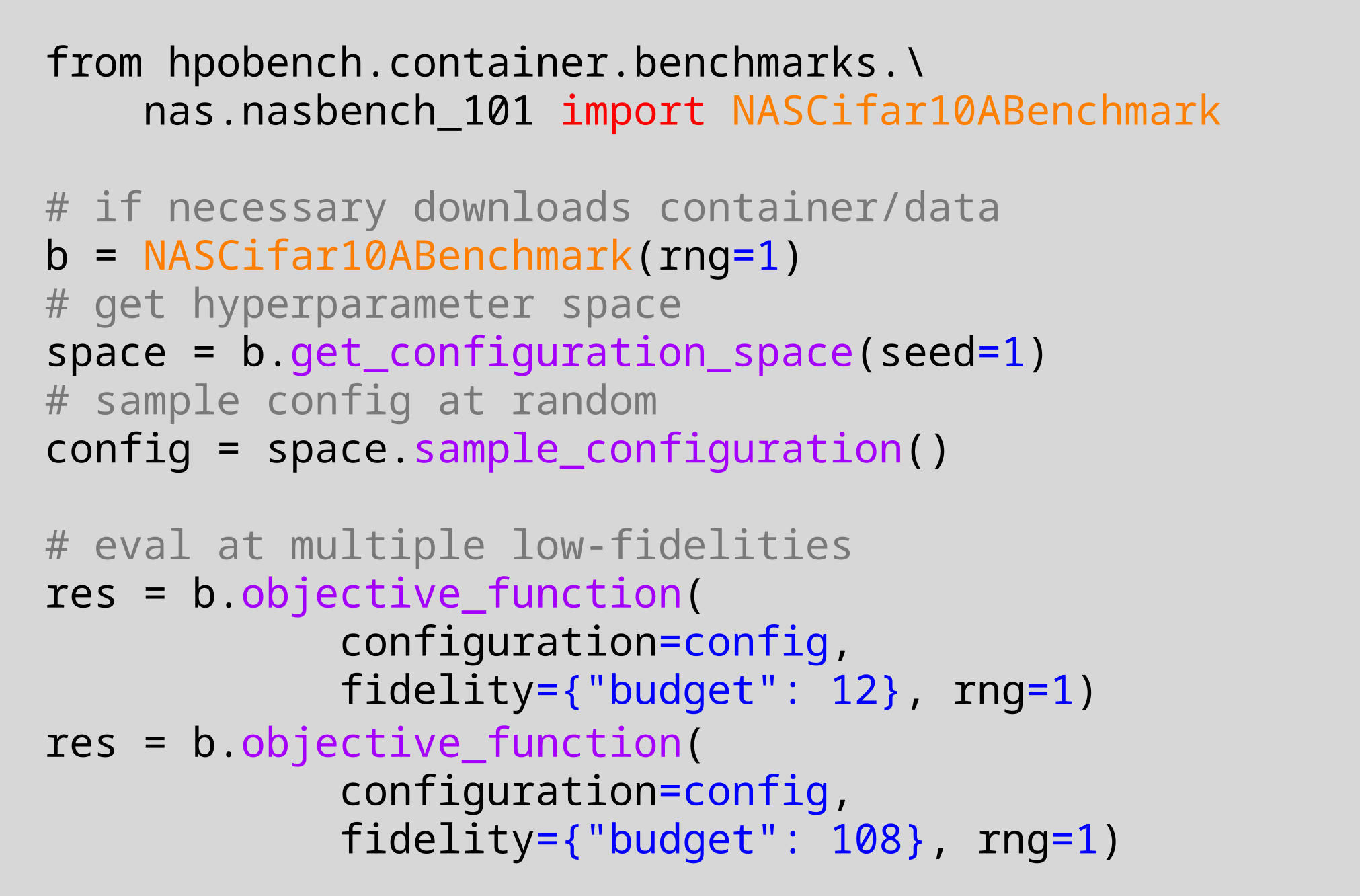}
    \caption{Code example initializing and evaluating a benchmark.}
    \label{fig:code}
\end{minipage}
\end{figure}

\hpobench{} circumvents such issues through the containerization of benchmarks using Singularity~\citep{kurtzer-plos17a} containers.\footnote{We chose Singularity over the popular Docker (\url{https://www.docker.com/}) alternative as it (1) does not require super user access and (2) is available on the computer clusters we have access to.} Each benchmark and its dependencies are packaged as a separate container, which isolates benchmarks from each other and also from the host system. Figure~\ref{fig:container} illustrates the advantages that containerization provides, especially when running multiple benchmarks in the same environment. Note that without containers, the environment needs to satisfy the union of all of its benchmarks' requirements (which may actually be mutually exclusive!), while with containers the dependencies for any given benchmark only need to be satisfied once: for the creation of the container. Importantly, the dependencies do not need to be satisfied again for using the benchmarks. Each benchmark is uploaded as a container to a GitLab container registry to provide the history of different versions of the benchmark. Hence, any benchmark created under the \hpobench{} paradigm remains usable without additional bookkeeping or installation overheads for long-term usage. Additionally, no effort is required for maintaining already containerized benchmarks, as long as the API does not change.
Although not recommended, each benchmark can also be installed locally along with its specific dependencies without using the containers. We provide a short code sample in Figure~\ref{fig:code}. 

Our notion of \emph{reproducibility} follows the Claerbout/Donoho/Peng convention as summarized by Barba~\citep{barba-arxiv18a}. We preserve benchmarks as containers, so that they can be used without installing all dependencies to obtain the same results. This does not immediately lead to \emph{replicability} on the level of the optimization results. Users need to make sure to for example run a sufficient number of seed replicates to avoid unstable results~\citep{bouthillier-icml19a} and to take hardware differences into account when comparing optimizer overhead. Our work differs from other efforts to provide reproducible research. We do not aim to make a single experiment reproducible as \emph{repo2docker}~\citep{forde-repro18a} and we also do not aim to package and distribute the whole runtime or workflows as \emph{Jupyter Notebooks}~\citep{kluyver-ppap16a} or R's \emph{knittr}~\citep{callahan-psbc16a}. 
%

\subsection{Flexibility}
\label{ssec:hpobench:flexibility}

\hpobench{} is a flexible framework that can be used to validate existing HPO research, and develop and improve HPO algorithms, with a focus on \mf{} methods. It consists of two sets of benchmarks, which we describe in turn: $\nbenchs$ existing \mf{} benchmarks from $\ncommunitybenchfams$ families that we collected from the \mf{} literature (Section \ref{sec:existing}); and 88 new benchmarks from $\nnewbenchfams$ families we created to allow a much more flexible use of \hpobench{} (Section \ref{sec:new}).

\begin{table}[tbp]
\centering
\small
\caption{Overview of raw (\yesmark), surrogate (\nomark) and tabular (\maybemark) benchmarks. We report the number of benchmarks per family (\emph{\#benchs}), the number of continuous (\emph{\#cont}), integer (\emph{\#int}), categorical (\emph{\#cat}), ordinal (\emph{\#ord}) hyperparameters and how many are log-scaled. Furthermore, we report the fidelity, the optimization budget and the number of configurations for tabular and surrogate benchmarks.
}
\label{tab:benchmarks}
\begin{tabular}{@{\hskip 0mm}
l@{\hskip 0mm}c
c@{\hskip 1mm}c@{\hskip 1mm}c@{\hskip 1mm}c
c@{\hskip 1mm}c@{\hskip 1mm}c@{\hskip 1mm}c
c
@{\hskip 0mm}}
\toprule
               Family & \#benchs & \#cont(log) & \#int(log) & \#cat & \#ord & fidelity & type & opt. budget & \#confs & Ref. \\
\midrule
\cartpole &  1 &  4(1) &   3(3) &   - &   - &   repetitions &  \yesmark & $1$d & - & \citep{falkner-icml18a} \\
\midrule{}
\pybnn{} &  2 &  3(1) &   2(2) &   - &   - & samples &  \yesmark & $1$d & - & \citep{falkner-icml18a} \\
\midrule{}
\paramnettime{} &    6 & 5 &   1 &   - &   - &   time & \nomark & $7$d & - & \citep{falkner-icml18a} \\
\midrule{}
\NASHPO{} &  4 & - &   - &   3 &   6 &  epochs & \maybemark & $10^7$sec & $62\ 208$ & \citep{klein-arxiv19a} \\
\midrule
\multirow{3}{*}{\NASOOO{}} & \multirow{3}{*}{3} & - &   - &  26 &   - &  \multirow{3}{*}{epochs} & \multirow{3}{*}{\maybemark} & \multirow{3}{*}{$10^7$sec} & \multirow{3}{*}{$423$k} & \multirow{3}{*}{\citep{ying-icml19a}} \\
{} & {} & - &   - &  14 &   - & {}  & {}  & {}  & {} & {}  \\
{} & {} & 21 &   1 &   5 &   - & {}  & {}  & {}  & {} & {}  \\
\midrule
\NASTOO{} & 3 &   - &   - &   6 &   - &  epochs & \maybemark & $10^7$sec & $15\ 625$ & \citep{dong-iclr20a} \\
\midrule
\multirow{3}{*}{\NASOSO{}} & \multirow{3}{*}{3} & - & - & 9 & - & \multirow{3}{*}{epochs} & \multirow{3}{*}{\maybemark} & \multirow{3}{*}{$10^7$sec} & $6\ 240$ & \multirow{3}{*}{\citep{zela-iclr20b}} \\
{} & {} & - & - & 9 & - & {} & {} & {} & $29\ 160$ & {} \\
{} & {} & - & - & 11 & - & {} & {} & {} & $363\ 648$ & {} \\
\midrule
\midrule
\lr{} & 20 & 2(2) & - & - & - & iter & \yesmark, \maybemark & $100 \times$ & 625 & \emph{new} \\
\svm{} & 20 & 2(2) & - & - & - & data & \yesmark, \maybemark & average & 441 & \emph{new} \\
\rf{} & 20 & 1 & 3(2) & - & - & \#trees & \yesmark, \maybemark & runtime on & 10k & \emph{new} \\
\xgb{} & 20 & 3(2) & 1(1) & - & - & \#trees & \yesmark, \maybemark & the highest & 10k & \emph{new} \\
\mlp{} & 8 & 2(2) & 3(2) & - & - & epochs & \yesmark, \maybemark & fidelity & 30k & \emph{new} \\
\bottomrule
\end{tabular}
\end{table}

\subsubsection{Existing Community Benchmarks} \label{sec:existing}

Firstly, to allow comparability with previous experiments, we collected $\nbenchs$ existing \mf{} benchmarks from $\ncommunitybenchfams$ families from the \mf{} literature; \hpobench{} preserves these benchmarks by containerizing them and encapsulating them all under a common API (which was not the case before). This not only ensures important previous work to remain accessible, but it also bypasses dependency issues enabling long term usage (see Appendix~\ref{app:dependencies}). 

Specifically, these benchmarks 
comprise raw benchmarks tuning a reinforcement learning agent (PPO on \cartpole{}~\citep{falkner-icml18a}) and a Bayesian neural network (\pybnn{}~\citep{falkner-icml18a}), a random forest-based surrogate benchmark tuning an MLP (\paramnettime{}~\citep{falkner-icml18a}) and four popular NAS benchmark families (\NASHPO{}~\citep{klein-arxiv19a}, \NASOOO{}~\citep{ying-icml19a}, \NASTOO{}~\citep{dong-iclr20a}, and \NASOSO{}~\citep{zela-iclr20b}).
However, these existing community benchmarks also have certain limitations: they are only of limited use for transfer HPO (since there are only between 1 and 6 benchmarks per family), they only offer a single fidelity dimension, and they only evaluate a single metric. We therefore augmented them with $\nnewbenchfams$ new families of benchmarks we describe next.

\subsubsection{New Benchmarks} \label{sec:new}

To substantially increase the range of possible applications of \hpobench{}, we defined $\nnewbenchfams$ new benchmark families with up to $20$ different datasets per family, comprising a total of $88$ new \mf{} benchmarks. These new benchmarks also provide multiple metrics and multiple fidelity dimensions to go beyond the aforementioned limitations of the community benchmarks. 

Our new benchmarks are based on the following popular ML algorithms: \textbf{\svm{}, \lr{}, \xgb{}, \rf{}}, and \textbf{\mlp{}}. All of them evaluate the respective ML algorithm as implemented in scikit-learn~\citep{scikit-learn} and XGBoost~\citep{chen-kdd16} on $20$ publicly available datasets ($8$ for the \mlp{} due to its high computational cost) from the OpenML AutoML benchmark~\citep{gijsbers-automl19a}.
We give the OpenML~\cite{vanschoren-sigkdd14a} task IDs in Table~\ref{tab:tids} in Appendix~\ref{app:benchmarks}, which provide fixed train-test splits; for each such task, we used $33\%$ of the training set as the validation split, determined through stratified sampling under a fixed seed. The entire objective function then consists of preprocessing, training the model on the remaining $67\%$ of the fixed OpenML training split, prediction on the fixed validation split, evaluating $4$ different metrics (see Appendix \ref{app:ssec:new}), and recording model fit and inference times.
The fidelities are algorithm-specific if possible (number of trees, iterations, epochs) or dataset subsets otherwise (which is used for \svm{}). These benchmarks are available both as raw and tabular versions, have the same API and exist in independent, non-conflicting containers; for the tabular versions, we discretized each hyperparameter (and fidelity) and evaluated $5$ different seeds for each configuration of the resulting grid.

Also, four of our new benchmark families (\lr{}, \rf{}, \xgb{}, \mlp{}) allow up to two fidelity dimensions. This enables the development and benchmarking of methods for \mf{} optimization with multiple fidelity dimensions, a direction that we deem very promising yet understudied. Similarly, our tabular data collected over multiple datasets (up to $20$) allows the effective use of these benchmarks for transfer-HPO, and the recording of multiple evaluation metrics also allows these benchmarks to be used for multi-objective optimization. Moreover, each configuration is recorded on different fidelities with their associated costs, which further lends \hpobench{} great potential in future research in cost-based meta-learning or multi-fidelity multi-objective optimization.

\begin{figure}[htbp]
    \centering
    \begin{tabular}{@{\hskip 0mm}
    c@{\hskip 0mm}c@{\hskip 0mm}
    c@{\hskip 0mm}c@{\hskip 0mm}
    c@{\hskip 0mm}c@{\hskip 0mm}}
    \lr{} & \rf{} & \svm{} & \xgb{} & \mlp{} \\
    \includegraphics[width=0.2\textwidth]{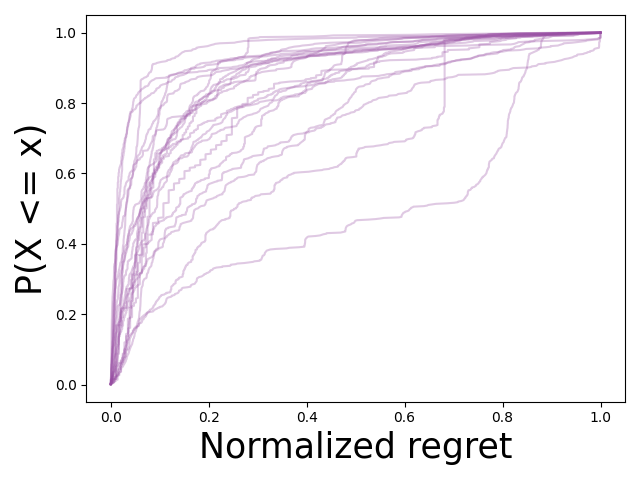} &
    \includegraphics[width=0.2\textwidth]{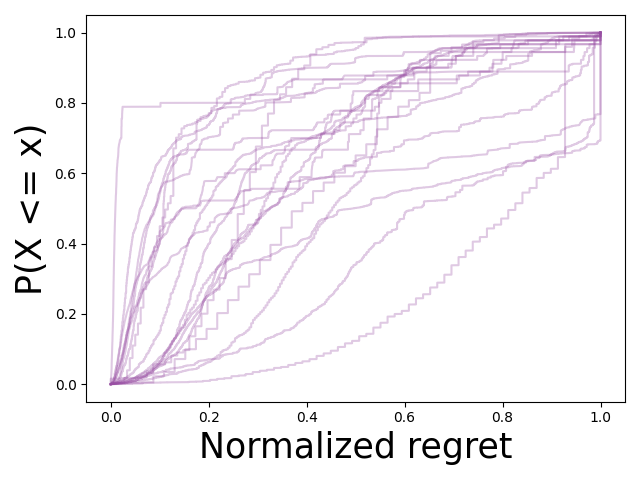} &
    \includegraphics[width=0.2\textwidth]{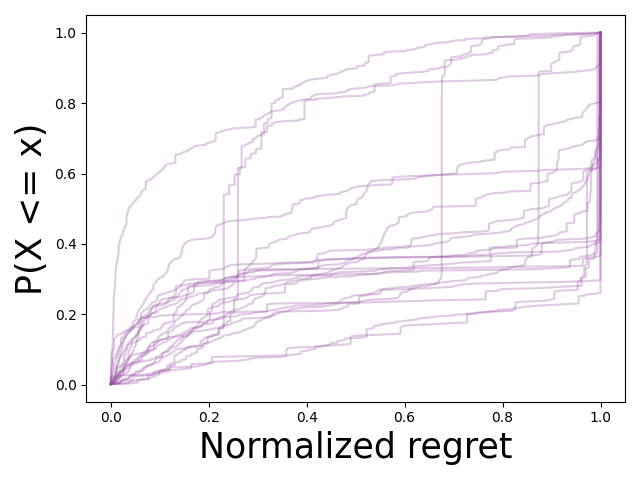} &
    \includegraphics[width=0.2\textwidth]{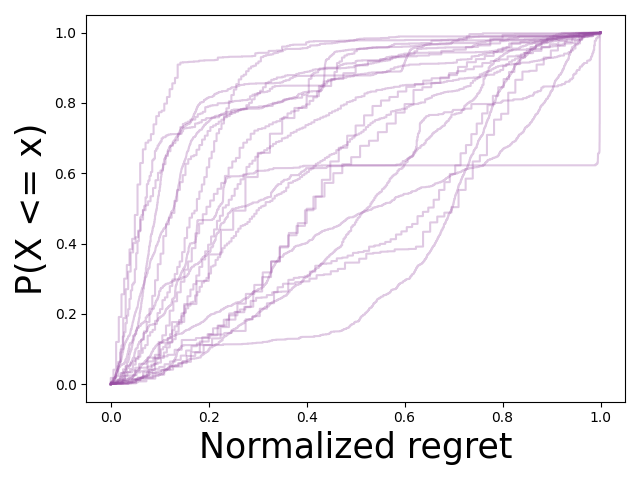} & \includegraphics[width=0.2\textwidth]{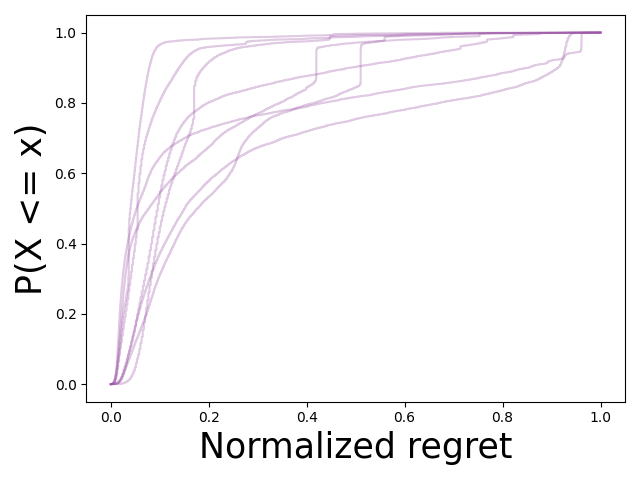} \\
    \end{tabular}
    \caption{Empirical cumulative distribution. Each plot corresponds to one ML algorithm, and each line within a plot corresponds to one dataset. The lines show the ECDF of the normalized regret of all evaluated configurations of the respective ML algorithm on the respective dataset.}
    \label{fig:ecdf}
\end{figure}

To demonstrate the diversity of our new benchmarks, we show the \emph{empirical cumulative distribution function} (ECDF) for each family in Figure~\ref{fig:ecdf}. Each line corresponds to one dataset and shows how the objective values are distributed. From the varying amounts of well and badly performing normalized regrets we can conclude that the benchmarks yield different landscapes and thus are diverse in smoothness, resulting in varying algorithm performance. 
Moreoever, the $\nnewbenchfams$ new spaces vary in their dimensionality (up to $5$ for \mlp{}), in the hyperparameter data types and their range (see Appendix~\ref{app:benchmarks}). 

%
\section{Experiments}
\label{sec:exp}

Now, we turn to an exemplary use of our benchmarks in order to demonstrate some features of \hpobench{} and its utility for HPO research. 
We used our benchmark suite to run a large-scale empirical study comparing $\nopts$ optimization methods on our $\nbenchfams$ benchmark families (we report detailed results in Appendix~\ref{app:moreresults}). We first give details on the experimental setup and then study the following two exemplary research questions: \textbf{(RQ1)} \emph{Do advanced methods improve over random baselines?} and \textbf{(RQ2)} \emph{Do \mf{} methods improve over single-fidelity methods?}

\subsection{Experimental Setup}
\label{ssec:exp:setup}

For each benchmark and optimizer, we conducted $32$ repetitions with different seeds to avoid reliance on individual seeds~\cite{bouthillier-icml19a}.
For our new benchmarks, which have multiple metrics, we minimized $1-$accuracy. For each run, we allowed an optimization budget as described in Table~\ref{tab:benchmarks} and accumulate time taken by the benchmark (recorded time for tabular benchmarks, predicted time for surrogate benchmarks and wallclock time for raw benchmarks; for our new benchmarks, we used the tabular versions to avoid unnecessary compute costs and CO2 exhaustion) and the optimizer (wallclock time).
We kept track of all evaluations and computed trajectories, i.e., the best-seen value at each time step, as follows: If for an evaluation we cannot find another evaluation conducted on the same or a higher fidelity, we treat it as the best-seen value; if it is on the highest fidelity evaluated so far, we treat it as the best seen value if it has a lower loss than the best-seen so far on that fidelity; otherwise, we do not consider this evaluation for the trajectory. This decision reflects the \mf{} setting, where a higher budget results in a better estimate of the actual value of interest but can cause jumps in the optimization trajectory, (e.g., when a configuration is the first to be evaluated on a higher budget but is worse than the best configuration on a lower budget). 
To aggregate and report results, we use either the \emph{final performance} (per benchmark, see Appendix~\ref{app:moreresults}), \emph{performance-over-time} (per benchmark, see Appendix~\ref{app:moreresults}) or \emph{rank-over-time} (across multiple benchmarks). For tabular and surrogate benchmarks we report optimization regret (the difference between the best-found value and the best-known value) and for the other benchmarks, we report the actual optimized objective value.\footnote{Since we study optimizers, we report optimization performance (in the case of ML the validation performance, which is the objective value seen by the optimizer. We note that \hpobench{} in principle allows to compute test performance (the loss computed on a separate test set on the highest fidelity).}

We give details on the hardware and required compute resources in Appendix~\ref{app:hardware} and~\ref{app:runtime} and release code for the experiments here: \url{https://github.com/automl/HPOBenchExperimentUtils}.

\subsection{Considered Optimizers}
\label{ssec:exp:opt}

We evaluated a wide set of optimizers including baselines for \bb{} and \mf{} optimization. Our selection of optimizers does \emph{not} aim at finding the best optimization algorithm, but to study a broad range of different implementations and tools (for more details see Appendix~\ref{app:opt}). 
As \bb{} optimizers, which only access the highest fidelity, we considered random search (\random{}), differential evolution (\de{}~\citep{storn-jgo1997a,awad-iclr20}) and \bo{} with different models: a Gaussian Process model (\smacbo{}~\cite{jones-jgo98a,lindauer-dso19a,lindauer-jmlr22a}), a random forest (\smacrf{}~\cite{hutter-lion11a}), a kernel density estimator (KDE) (\tpe{}~\citep{falkner-icml18a}). Lastly, we also used the winning solution of the NeurIPS BBO challenge (\hebo{}~\citep{cowenrivers-jair22a,turner-neuripscomp20a}). 
For \mf{} optimization, we used \mf{} extensions of some methods mentioned above: Hyperband (\hb{}~\cite{li-jmlr18a}) and its combination with KDE-based \bo{} (\bohb{}~\citep{falkner-icml18a}), with RF-based \bo{} (\smachb{}~\citep{lindauer-jmlr22a}) and with \de{} (\dehb{}~\citep{awad-ijcai21}). 
Additionally, we use Dragonfly~\citep{kandasamy-jmlr20a}
using a GP with \mf{} optimization and combinations of optimization and \mf{} algorithms implemented in 
Optuna~\citep{akiba-kdd19a} (see appendix).~\footnote{We include this framework to show compatibility of \hpobench{} with popular frameworks, but note that it expects to freeze and thaw evaluations. \hpobench{} implements a stateless objective function and, thus, runs that could be thawed and continued instead get accounted the full costs of rerunning them, which slows down optimization. We defer stateful benchmarks to future work.}

\subsection{RQ1: Do advanced methods improve over random search?}
\label{ssec:exp:imp_over_random}

To demonstrate the validity of our benchmarks, we independently replicate the findings of the 1st NeurIPS Blackbox Optimization challenge~\citep{turner-neuripscomp20a}: ``\emph{decisively showing that \bo{} and similar methods are superior choices over \random{} and grid search for tuning hyperparameters of ML models}''. While this question has already been studied before~\citep{bergstra-nips11a,bergstra-jmlr12a,bergstra-icml13a,thornton-kdd13a,wistuba-ml18a,eriksson-neurips19a,balandat-neurips20a}, we will also study it \wrt{}\mf{} optimization and using the popular \hb{} baseline. We leave out grid search as \random{} has been shown to be superior~\citep{bergstra-jmlr12a} and as there is no \mf{} version of grid search.

\begin{figure}[htbp]
\small
    \centering
    \begin{tabular}{@{\hskip 0cm}c@{\hskip 0cm}c@{\hskip 0cm}c@{\hskip 2mm}|@{\hskip 0mm}c@{\hskip 0cm}}
    \bb{} & \mf{} & \bb{} + \mf{} & subsets \\
\includegraphics[width=0.245\textwidth]{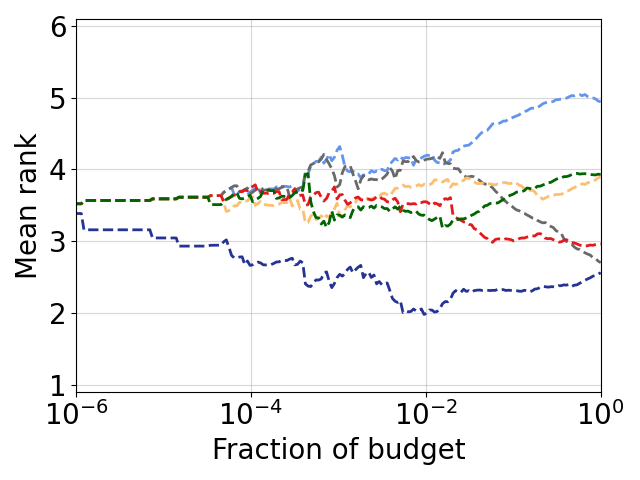} &
\includegraphics[width=0.245\textwidth]{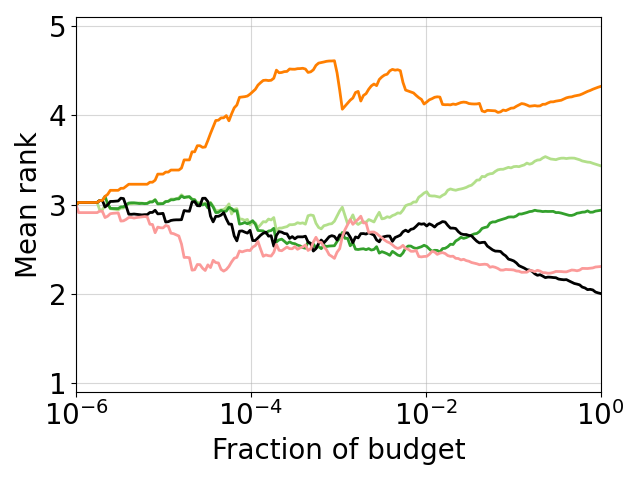} &
\includegraphics[width=0.245\textwidth]{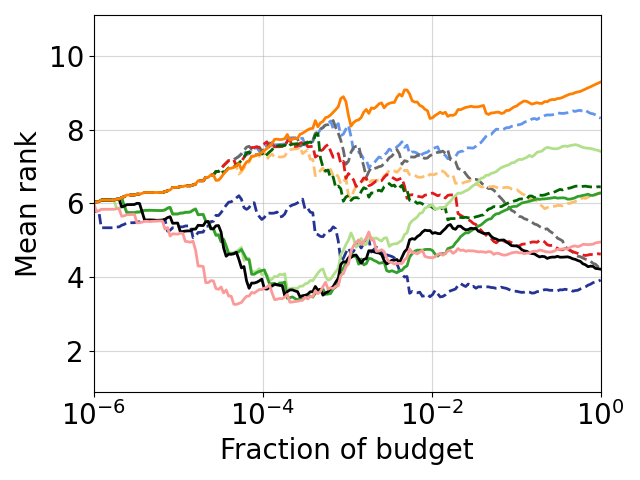} &
\includegraphics[width=0.245\textwidth]{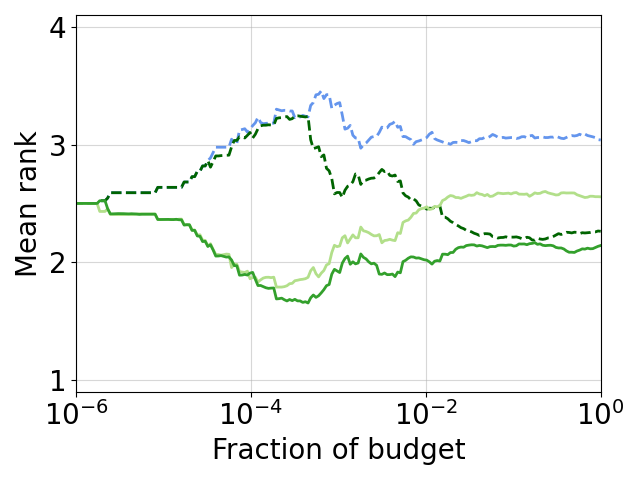} \\
\includegraphics[width=0.245\textwidth]{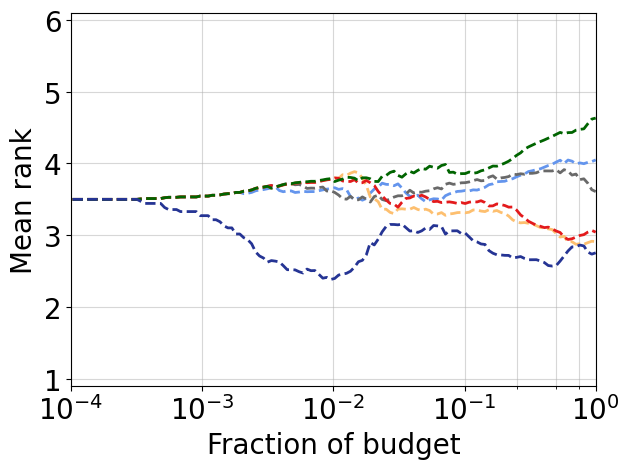} &
\includegraphics[width=0.245\textwidth]{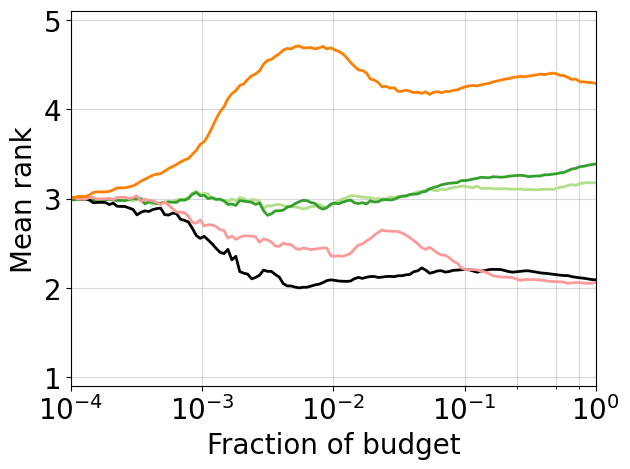} &
\includegraphics[width=0.245\textwidth]{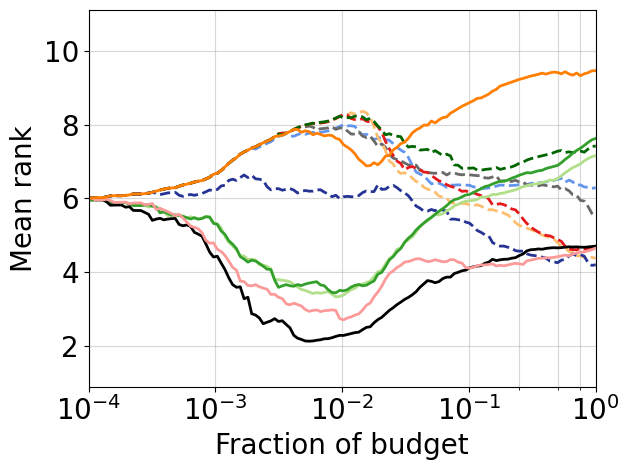} &
\includegraphics[width=0.245\textwidth]{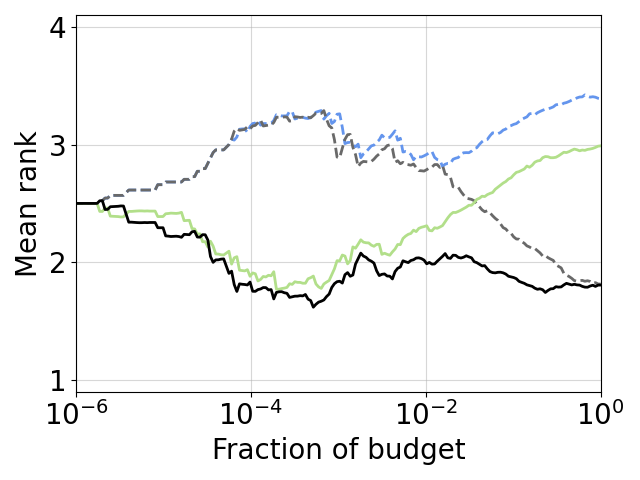} \\
        \end{tabular}
\includegraphics[width=0.9\textwidth]{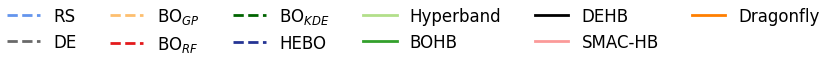} 
\caption{Mean \emph{rank-over-time} across $32$ repetitions of different sets of optimizers (lower is better). The left part shows rank across all \emph{existing community} (upper row) and \emph{new} (lower row) benchmarks . The right part reports results on the \emph{existing community} benchmarks only for subsets of optimizers.}
\label{fig:ranks}
\end{figure}

We report ranks-over-time in Figure~\ref{fig:ranks}, comparing \bb{} (\de{}, \smacbo{}, \smacrf{}, \hebo{}, \tpe{}; 1st column) and \mf{} (\bohb{}, \dehb{}, \smachb{}, \dragonfly{}; 2nd column) optimizers on \emph{existing community} (top row) and \emph{new} (bottom row) benchmarks. On both benchmark sets, most \bb{} and \mf{} optimizers clearly outperform the respective baseline (\random{} (blue) and \hb{} (light green)) on average. 
We also observe that \bo{} improves over the evolutionary algorithm \de{} in the beginning, but, except for \hebo{}, looses to it in the very long run on the \emph{existing community} benchmarks~\citep{hutter-arXiv13a,klein-neurips19a,awad-ijcai21}. This does not happen on the \emph{new} benchmarks, as their time limits are set more aggressively and the methods developed for this setting (\hebo{}~\citep{cowenrivers-jair22a}, \smacbo{}, \smacrf{} and \smachb{}~\citep{lindauer-dso19a}) achieve lower ranks. 
Considering per-benchmark results (Appendix~\ref{app:moreresults}), we also observe that methods which appear clearly inferior in the ranking plots perform very well on individual benchmarks (e.g. \dragonfly{}\footnote{We would like to note that the bad rank of \dragonfly{} for some benchmarks is due to its overhead which prevented it from spending sufficient budget on function evaluations; see Section~\ref{app:runtime} for details.} on \NASTOO{}). Finally, we find \hebo{} to substantially improve over all other \bb{} methods.

Besides qualitative measures, we also quantitatively measure whether the advanced methods outperform the respective baselines by counting the number of wins, ties and losses and using the sign test to verify significance~\citep{demsar-06a} on the existing community benchmarks in Table~\ref{tab:pvalues-rq1} (the new benchmarks yield similar results; see Appendix~\ref{app:moreresults}). We can observe that four out of five \bb{} methods are significantly better than \random{}. In the \mf{} case, only two out of four methods are significantly better than \hb{} and one method is consistently worse than \hb{}. Overall, we conclude that advanced methods consistently outperform random search.
\begin{table}[htbp]
    \centering
    \small
    \caption{P-value of a sign test for the hypothesis that advanced methods outperform the baseline \random{} for \bb{} optimization and \hb{} for \mf{} optimization. We underline p-values that are below $\alpha = 0.05$ and boldface p-values that are below $\alpha = 0.05$ after multiple comparison correction (dividing $\alpha$ by the number of comparisons, i.e. 5 and 4; boldface/underlined implies that the advanced method is better than \random{}/\hb{}). We also give the wins/ties/losses of \random{} and \hb{} against the challengers.}
    \label{tab:pvalues-rq1}
    \begin{tabular}{lrrrrr}
        \toprule
        {} & \de & \smacbo & \smacrf & \hebo{} & \tpe{}  \\ 
        p-value against \random{} & $\mathbf{\underline{0.00043}}$ & $\underline{0.01330}$ & $\mathbf{\underline{0.00001}}$ & $\mathbf{\underline{0.00217}}$ & $0.06690$  \\
        wins/ties/losses against \random{} & $18/2/2$ & $15/3/4$ & $19/3/0$ & $17/2/3$ & $13/4/5$ \\
        \midrule
        {} & \bohb & \dehb & \smachb & \dragonfly & {} \\
        p-value against \hb{} & $0.06690$ & $\mathbf{\underline{0.00001}}$ & $\mathbf{\underline{0.00011}}$ & $0.99783$ & {} \\
        wins/ties/losses against \hb{} & $13/4/5$ & $20/2/0$ & $18/3/1$ & $5/0/17$ & {} \\
        \bottomrule
    \end{tabular}
\end{table}
%
\subsection{RQ2: Do \mf{} methods improve over \bb{} methods?}
\label{ssec:exp:fidelities}

Next, we study whether \mf{} optimization methods are able to consistently improve over \bb{} optimization methods given a fixed time budget. 
For this, we look again at ranking-over-time in Figure~\ref{fig:ranks}. We first compare \bb{} methods with their respective \mf{} extension, i.e., \de{} vs. \dehb{} and \tpe{} vs. \bohb{} in the two plots in the rightmost column. We can see that in the beginning \hb{} and the \mf{} optimizers perform very similarly and consistently outperform \random{} and the respective \bb{} version. After a while, the \mf{} versions improve over the \hb{} baseline, and given enough time, the \bb{} versions catch up. Second, we compare all optimizers on the \emph{existing community} (3rd column, top) and \emph{new} (3rd column, bottom) benchmarks. Here, we can observe a similar pattern in that \hb{} is a very competitive baseline in the beginning but is outperformed first by the advanced \mf{} methods and then also by the \bb{} methods. This is less pronounced on the \emph{new} benchmarks, which we attribute to the tighter time limits. 

Similarly to RQ1, we counted wins, ties and losses and used the sign test to verify significance~\citep{demsar-06a} on the community benchmarks for $100\%$, $10\%$ and $1\%$ of the total budget in Table~\ref{tab:pvalues-rq2} (the new benchmarks yield similar results; see Appendix~\ref{app:moreresults}). We can observe that only \hb{} is able to outperform its \bb{} counterpart for all three budgets we check. For three \mf{} methods there is a significant improvement over the \bb{} methods for $1\%$ of the budget. 
For the full budget we can no longer state that any of the advanced \mf{} methods is statistically better than their counterpart, but 
judging by the wins and losses, the advanced \mf{} methods are still competitive.

Overall, \mf{} optimizers outperform \bb{} optimizers for relatively small compute budgets. Given enough budget, \bb{} optimizers become competitive with their \mf{} versions; in particular, \de{} and \smacrf{} performed very well in the end. However, we need to take into account that for the \emph{existing community} benchmarks the potential catch-up (if at all) only happens after a very substantial amount of (simulated) wallclock time (e.g., 10 Mio.\ seconds). Hence, \mf{} methods are crucial to efficiently tackle real, expensive optimization problems.
\begin{table}[htbp]
    \small
    \centering
    \caption{P-values of a sign test for the hypothesis that \mf{} outperform their \bb{} counterparts. We boldface p-values that are below $\alpha = 0.05$ (implying that \mf{} is better).}
    \label{tab:pvalues-rq2}
    \begin{tabular}{lrrrrr}
        \toprule
        Budget & {} & \hb{} vs \random{} & \dehb{} vs \de{} & \bohb{} vs \tpe{} & \smachb{} vs \smacrf{} \\
        \midrule
        \multirow{2}{*}{$100\%$} & p-values & $\mathbf{0.00074}$ & $0.73827$ & $0.14314$ & $0.73827$ \\ 
        & w/t/l & $16/5/1$ & $6/8/8$ & $12/4/6$ & $6/8/8$ \\
        \midrule
        \multirow{2}{*}{$10\%$} & p-values & $\mathbf{0.00845}$ & $0.09462$ & $0.14314$ & $0.50000$ \\
         & w/t/l & $16/2/4$ & $10/9/3$ & $12/4/6$ & $8/7/7$ \\
        \midrule
        \multirow{2}{*}{$1\%$} & p-values & $\mathbf{0.00074}$ & $\mathbf{0.03918}$ & $0.06690$ & $\mathbf{0.03918}$ \\
        & w/t/l & $17/3/2$ & $14/3/5$ & $14/2/6$ & $13/5/4$ \\
        \bottomrule
    \end{tabular}
\end{table}

To conclude, in general when low-fidelities are available and they are representative of the true objective function, \mf{} methods are clearly beneficial. In practice, we found that \dehb{} and \smachb{} are reliable \mf{} optimizers that work well across the whole collection of benchmarks, while other \mf{} optimizers are not able to improve over \hb{} consistently. By exploring a very broad range of benchmarks, we also found an existence proof that \bb{} methods can outperform \mf{} methods for very high budgets and that even advanced methods can be outperformed by \random{} in individual benchmarks. We pose it as a challenge to the field 
to develop methods that do not exhibit poor performance in \emph{any} of the many benchmarks in \hpobench{}.

\section{Discussion and Future Work}
\label{sec:discussion}

We proposed \hpobench{}, a library for \mf{} HPO benchmarks. It serves two purposes: (a) to provide benchmarks with a unified API, and (b) to make them easy to install and use by containerizing them and thus enable rapid prototyping and the development of new \mf{} methods that are crucial for ML research and applications. Finally, our library is open-source and we welcome contributions of new benchmarks to keep the library up-to-date and evolve it.

On the technical side, so far, we focused on developing a benchmark library, but we see a large potential in connecting our library with other benchmarking frameworks (e.g.  \href{https://github.com/numbbo/coco}{COCO}~\citep{hansen-oms20} and \href{https://github.com/uber/bayesmark}{Bayesmark}~\cite{bayesmark}), optimization frameworks (e.g. \href{https://github.com/facebookresearch/nevergrad}{Nevergrad}~\citep{nevergrad} and \href{https://github.com/sherpa-ai/sherpa}{Sherpa}~\citep{hertel-softwarex20a}) and extending it with further benchmarks~\citep{bliek-arxiv21a,haese-MLST21,klein-neurips19a,siems-arxiv20a,xiao-arxiv21a,sehic-arxiv21a,pfisterer-arxiv21a} to increase diversity and to simplify evaluation and comparison of optimizers. For this, it would be interesting to also containerize the optimizers since they can suffer from the same issues as benchmarks.
Furthermore, so far, \hpobench{} only contains stateless benchmarks starting a single container. We would like to extend the library to also support optimizers requiring stateful benchmarks (to freeze and thaw evaluations) or running in parallel.

Our set of benchmarks already covers raw, tabular, and surrogate benchmarks, but it would be useful to have all three versions available for all benchmarks, and to automatically generate tabular and surrogate-based benchmarks from raw benchmarks. Also, our new benchmarks can be used to evaluate multi-objective (multiple metrics) and meta-learning (across datasets) methods or even meta-learned \mf{} multi-objective methods. We hope for the community to play a large role in defining the protocols for the different special cases; e.g., budgets need to be set differently for \bb{} multi-objective optimization and single-objective hyperparameter transfer learning. Additionally, it would be interesting to study hierarchical search spaces to cover work in \automl{}. Furthermore, there is a large potential in automatically creating \mf{} benchmarks from any ML algorithm by using data subsets as a low-fidelity. 

We also conducted a large-scale study evaluating $\nopts$ algorithm implementations to demonstrate compatibility with a wide range of optimization tools, and we thus believe that our library is well suited for future research on \mf{} optimization. We showed that advanced HPO methods are preferable over \random{} and \hb{} baselines, and that \mf{} extensions of popular optimizers improve over their \bb{} version.
Lastly, to reduce computational effort, we would like to study whether we can learn which benchmarks are hard and whether there is a representative subset of them~\citep{cardoso-arxiv2021a}.

\begin{ack}
We would like to thank Stefan Stäglich and Archit Bansal for code contributions and Stefan Falkner for useful discussions and comments on an early draft of this project. This work has partly been supported by the European Research Council (ERC) under the European Union’s Horizon 2020 research and innovation programme under grant no. 716721, and by TAILOR, a project funded by EU Horizon 2020 research and innovation programme under GA No 952215. Robert Bosch GmbH is acknowledged for financial support. The authors also acknowledge support by the state of Baden-W\"{u}rttemberg through bwHPC and the German Research Foundation (DFG) through grant no INST 39/963-1 FUGG.
\end{ack}

\bibliography{strings,lib,local,proc}
\bibliographystyle{unsrtnat}

\newpage
\appendix

\section{Maintenance}

In this section we present a maintenance plan that is adapted from the datasheets for datasets~\citep{gebru-arxiv20a}.

\begin{itemize}
    \item{\textbf{Who is maintaining the benchmarking library?}} \hpobench{} is developed and maintained by the Machine Learning Lab at the University of Freiburg. 
    \item{\textbf{How can the maintainer of the dataset be contacted(e.g., email address)?}}
    Questions should be submitted via an issue on the Github repository at \url{https://github.com/automl/HPOBench}.
    \item{\textbf{Is there an erratum?}}
    No.
    \item{\textbf{Will the benchmarking library be updated?}}
    We consider adding new benchmarking problems and potentially fix existing issues with existing benchmarks. Such changes will be communicated via release notes in Github releases. 
    \item{\textbf{Will older versions of the benchmarking library continue to be supported/hosted/maintained?}}
    Older versions of the benchmarking code are available via the underlying git repository. Containers are versioned and available via Gitlab. We aim to answer questions on a best-effort basis, but will not do so for older versions of the benchmarking library.
    \item{\textbf{If others want to extend/augment/build on/contribute to the dataset, is there a mechanism for them to do so?}}
    We allow contributions from the community via a process that is currently described at \url{https://github.com/automl/HPOBench/wiki/How-to-add-a-new-benchmark-step-by-step}.
    \item{\textbf{Any other comments?}}
    No.
\end{itemize}

\section{Benchmarking efforts}
\label{app:related:otherframeworks}
In addition to Section~\ref{sec:related} of the main paper, we provide here a non-exhaustive list of further benchmarking libraries in the area of HPO consisting of not only a publication but which constitute or constituted a long-running effort to compare methods:

\begin{itemize}
    \item \href{https://github.com/automl/HPOlib}{HPOLib}~\citep{eggensperger-bayesopt13} to benchmark global optimization methods
    \item \href{https://bitbucket.org/mlindauer/aclib2}{ACLib}~\citep{hutter-lion14a} to benchmark algorithm configuration methods
    \item \href{https://gym.openai.com/}{OpenAI Gym}~\citep{brockman-arxiv16a} to benchmark RL methods
    \item \href{https://github.com/numbbo/coco}{COCO}~\citep{hansen-oms20} to compare continuous optimization methods
    \item \href{https://github.com/uber/bayesmark}{Bayesmark}~\citep{bayesmark} to benchmark Bayesian optimization methods
    \item \href{https://www.openml.org/s/99}{OpenML benchmarking suites}~\citep{bischl-neurips21a} provide a set of datasets for supervised classification
    \item \href{https://github.com/aspuru-guzik-group/olympus}{Olympus}~\citep{haese-MLST21} provides a set of experiment planning tasks to evaluate optimization algorithms
    \item \href{https://github.com/releaunifreiburg/HPO-B}{HPO-B}~\citep{pineda-neurips21a} provides a tabular benchmark to compare \bb{} HPO methods
    \item \href{https://github.com/AlgTUDelft/ExpensiveOptimBenchmark}{ExpoBench}~\citep{bliek-arxiv21a} provides expensive benchmark problems for HPO
\end{itemize}

\section{Benchmarking competitions}
\label{app:related:competitions}

In addition to Section~\ref{sec:related} of the main paper, we provide here a non-exhaustive list of benchmarking competitions on HPO and related topics:
\begin{itemize}
    \item The AutoML challenges~\cite{guyon-automl19a}
    \item The AutoDL challenge~\cite{liu-hal20a}
    \item NeurIPS 2020 Black-Box optimization challenge~\citep{turner-neuripscomp20a}
    \item The KDD cup (see \url{https://www.kdd.org/kdd-cup})
    \item Challenges in Machine Learning (CIML) workshop series (see \url{https://ciml.chalearn.org/})
    \item Black-box Optimization Benchmarking (BBOB) workshop series~\citep{molina-cc18a} (see \url{https://numbbo.github.io/workshops/})
\end{itemize}

\section{More Details on Considered Benchmarks}
\label{app:benchmarks}

In addition to the main paper, here we provide further details on our benchmarks collected. We start with issues we faced during collection and then briefly describe the existing community benchmarks (Section~\ref{app:ssec:existing}) and the new benchmarks (Section~\ref{app:ssec:new}).

\subsection{Conflicting Dependencies.}
\label{app:dependencies}

During benchmark collection, we also encountered a few examples of conflicting dependencies and updated interfaces making long-term maintenance of non-containerized benchmarks hard: \paramnettime{}~\citep{falkner-icml18a} was built with the latest version of scikit-learn~\citep{scikit-learn} (0.18) when it was developed but is incompatible with the current version (0.24); the \cartpole{} benchmark does not run with the latest version of TensorForce~\citep{kuhnle-web17a} due to a change in the API; \NASTOO{}~\citep{dong-iclr20a} changed its interface as well as the underlying data from its initial release. Additionally, in total, none of the \emph{existing community} benchmarks we collected for this paper had a full list of dependencies given.

\subsection{Existing Community Benchmarks}
\label{app:ssec:existing}

Here, we provide more details on the \emph{existing community} benchmarks currently in \hpobench{} and list their hyperparameter and fidelity spaces in Table~\ref{app:tab:space:exist}.

\begin{table}[p]
\small
\centering
\caption{Hyperparameter spaces of our benchmarks. For each benchmark, we report the hyperparameter names, type, whether they are on a log scale, and their respective range for each benchmark. Additionally, we report the same information for the fidelity space. If the spaces are different for different benchmarks within one family, we report them separately.}
\label{app:tab:space:exist}
\begin{tabular}{@{\hskip 0mm}llllr@{\hskip 0mm}}
        \toprule
        benchmark &  name & type & log & range \\
        \midrule
\multirow{7}{*}{\cartpole{}} 
          & batch\_size & int &   \yesmark &      $[8, 256]$ \\
          &    discount &  float &  \nomark &    $[0.0, 1.0]$ \\
          &   entropy\_regularization &  float &  \nomark &    $[0.0, 1.0]$ \\
          &   learning\_rate &  float &   \yesmark &  $[1e^{-07}, 0.1]$ \\
          & likelihood\_ratio\_clipping &  float &  \nomark &    $[1e^{-7}, 1.0]$ \\
          &  n\_units\_\{1,2\}* &    int &   \yesmark &      $[8, 128]$ \\
\cmidrule{2-5}
          &  repetitions &    int &   \nomark &      $[1, 9]$ \\
\midrule
\multirow{5}{*}{\pybnn{}}
          & burn\_in &  float &  \nomark &    $[0.0, 0.8]$ \\
          & l\_rate &  float &   \yesmark &  $[1e^{-6}, 0.1]$ \\
          &  mdecay &  float &  \nomark &    $[0.0, 1.0]$ \\
          & n\_units\_\{1,2\}* &    int &   \yesmark &     $[16, 512]$ \\
\cmidrule{2-5}
          &  epochs &    int &   \nomark &      $[500, 10000]$ \\
\midrule
\multirow{6}{*}{\paramnettime{}}
          & average\_units\_per\_layer\_log2 &  float &  \nomark &    [4.0, 8.0] \\
          & batch\_size\_log2 &  float &  \nomark &    $[3.0, 8.0]$ \\
          & dropout &  float &  \nomark &    $[0.0, 0.5]$ \\
          & final\_lr\_fraction\_log2 &  float &  \nomark &   $[-4.0, 0.0]$ \\
          & initial\_lr\_log10 &  float &  \nomark &  $[-6.0, -2.0]$ \\
          & num\_layers &    int &  \nomark &        $[1, 5]$ \\
\cmidrule{2-5}
\multicolumn{1}{l}{\textit{adult, higgs, mnist}} &  \multirow{4}{*}{epochs} & int &  \nomark & $[9, 243]$ \\
          \multicolumn{1}{l}{\textit{letter}} &   &  int & \nomark & $[3, 81]$ \\
          \multicolumn{1}{l}{\textit{optdigits}} &   &  int & \nomark & $[1, 27]$ \\
          \multicolumn{1}{l}{\textit{poker}} &   &  int & \nomark & $[81, 2187]$ \\
        \midrule
          \multirow{7}{*}{\NASHPO{}}
          & activation\_fn\_\{1, 2\}* &  cat &  - &                             \{tanh, relu\} \\
          & batch\_size &  ord &  - &                          \{8, 16, 32, 64\} \\
          & dropout\_\{1, 2\}* &  ord &  - &                          \{0.0, 0.3, 0.6\} \\
          & init\_lr &  ord &  - &  \{0.0005, 0.001, 0.005, 0.01, 0.05, 0.1\} \\
          & lr\_schedule &  cat &  - &                          \{cosine, const\} \\
          & n\_units\_\{1, 2\}* &  ord &  - &              \{16, 32, 64, 128, 256, 512\} \\
        \cmidrule{2-5}
          &  epochs &    int &   \nomark &      [3, 100] \\
        \midrule
          \multirow{4}{*}{\NASTOO{}}
            & 1<-0 &  cat &  - &  \{none, skip\_connect, \\
            & 2<-\{0,1\}$^*$ &  cat &  - &  nor\_conv\_1x1, nor\_conv\_3x3, \\
            & 3<-\{0,1,2\}$^*$ &  cat &  - & avg\_pool\_3x3\} \\
        \cmidrule{2-5}
          &  epochs &    int &   \nomark &      [12, 200] \\
        \midrule 
          \multirow{4}{*}{\NASA{}}
          &  edge\_\{0, 1, ..., 20\}* &  cat &  - &                                          \{0, 1\} \\
          & op\_node\_\{0, 1, .., 4\}* &  cat &  - &  \{conv1x1-bn-relu, conv3x3-bn-relu, \\
          & & & & maxpool3x3\} \\
        \cmidrule{2-5}
          &  epochs &    ord & \nomark &\{[4, 12, 36, 108\} \\
        \midrule 
          \multirow{3}{*}{\NASB{}}
            &  edge\_\{0, 1, ..., 8\}* &  cat &  - & \{0, 1, 2, ..., 20\} \\
            & op\_node\_\{0, 1, ..., 4\}* &  cat &  - &  \{conv1x1-bn-relu, conv3x3-bn-relu, \\
          & & & & maxpool3x3\} \\
        \cmidrule{2-5}
          &  epochs &    ord & \nomark & \{4, 12, 36, 108\} \\
        \midrule 
          \multirow{4}{*}{\NASC{}}
           & edge\_\{0, 1, ..., 20\}* &  float & \nomark &                                      [0.0, 1.0] \\
           & num\_edges &    int & \nomark &                                          [0, 9] \\
           & op\_node\_\{0, 1, ..., 4\}* &    cat & - &  \{conv1x1-bn-relu, conv3x3-bn-relu, \\
          & & & & maxpool3x3\} \\
        \cmidrule{2-5}
          &  epochs &    ord & \nomark & \{4, 12, 36, 108\} \\
        \bottomrule
    \end{tabular}
\end{table}
\begin{table}[t]
\small
\centering
\caption{Table~\ref{app:tab:space:exist} continued}
\begin{tabular}{@{\hskip 0mm}llllr@{\hskip 0mm}}
\toprule
\multirow{8}{*}{\NASOSOA{}}
    & choice\_block\_\{1,2,3,4\}\_op* & cat & - & \{conv1x1-bn-relu, conv3x3-bn-relu, \\
    & & & & maxpool3x3\} \\
    & choice\_block\_1\_parents & cat & - & \{(0,)\} \\
    & choice\_block\_2\_parents & cat & - & \{(0,1)\} \\
    & choice\_block\_3\_parents & cat & - & \{(0,1), (0,2), (1,2)\} \\
    & choice\_block\_4\_parents & cat & - & \{(0,1), (0,2), (0,3), (1,2), (1,3), (2,3)\} \\
    & choice\_block\_5\_parents & cat & - & \{(0,1), (0,2), (0,3), (0,4), (1,2), (1,3), \\
    & & & & (2,3), (1,4), (2,3), (2,4), (3,4)\} \\
    \cmidrule{2-5}
          &  epochs &    ord & \nomark & \{4, 12, 36, 108\} \\
    \midrule
\multirow{8}{*}{\NASOSOB{}}
      & choice\_block\_\{1,2,3,4\}\_op* &  cat &  - &  \{conv1x1-bn-relu, conv3x3-bn-relu, \\
      & & & & maxpool3x3\} \\
      & choice\_block\_1\_parents &  cat &  - & \{(0,)\} \\
      & choice\_block\_2\_parents &  cat &  - & \{(0,), (1,)\} \\
      & choice\_block\_3\_parents &  cat &  - & \{(0, 1), (0, 2), (1, 2)\} \\
      & choice\_block\_4\_parents &  cat &  - & \{(0, 1), (0, 2), (0, 3), (1, 2), (1, 3), (2, 3)\} \\
      & choice\_block\_5\_parents &  cat &  - &  \{(0, 1, 2), (0, 1, 3), (0, 1, 4), (0, 2, 3), (0, 2, 4), \\
      & & & & (0, 3, 4), (1, 2, 3), (1, 2, 4), (1, 3, 4), (2, 3, 4)\}\\
 \cmidrule{2-5}
     & epochs & ord & \nomark & \{4, 12, 36, 108\} \\
    \midrule
\multirow{12}{*}{\NASOSOC{}}
      & choice\_block\_\{1,2,3,4,5\}\_op* &  cat &  - & \{conv1x1-bn-relu, conv3x3-bn-relu, \\
      & & & & maxpool3x3\} \\
      & choice\_block\_1\_parents &  cat &  - & {(0,)} \\
      & choice\_block\_2\_parents &  cat &  - & \{(0,), (1,)\} \\
      & choice\_block\_3\_parents &  cat &  - & \{(0,), (1,), (2,)\} \\
      & choice\_block\_4\_parents &  cat &  - & \{(0, 1), (0, 2), (0, 3), (1, 2), (1, 3), (2, 3)\} \\
      & choice\_block\_5\_parents &  cat &  - & \{(0, 1), (0, 2), (0, 3), (0, 4), (1, 2), (1, 3),\\
      & & & & (1, 4), (2, 3), (2, 4), (3, 4)\} \\
      & choice\_block\_6\_parents &  cat &  - &  \{(0, 1), (0, 2), (0, 3), (0, 4), (0, 5), (1, 2),\\
      & & & & (1, 3), (1, 4), (1, 5), (2, 3), (2, 4), (2, 5), \\
      & & & & (3, 4), (3, 5), (4, 5)\} \\
 \cmidrule{2-5}
     & epochs & ord & \nomark & \{4, 12, 36, 108\} \\
    \bottomrule
    \end{tabular}
\end{table}

\noindent\textbf{\cartpole{}}~\citep{falkner-icml18a} A highly stochastic benchmark having $7$ hyperparameters of the \emph{proximal policy optimization}~\citep{schulman-arxiv17a} algorithm implemented in TensorForce~\citep{kuhnle-web17a} for the \emph{cartpole swing-up} task implemented in the OpenAI Gym~\citep{brockman-arxiv16a}. The number of repetitions is used as the fidelity and this benchmark is available only as a \emph{raw} benchmark.

\noindent\textbf{\pybnn{}}~\citep{falkner-icml18a} The Bayesian neural network benchmark is a $4$-hyperparameter tuning task to minimize the negative log-likelihood of a Bayesian neural network trained with stochastic gradient Hamilton Monte-Carlo~\citep{chen-icml14} with scale adaption~\citep{springenberg-nips16a} on two different regression datasets from the UCI repository (\cite{uci-19}, Protein Structure and YearPredictionMSD). It is implemented with Lasagne~\citep{lasagne-15a} and Theano~\citep{theano-arXiv16a}. It uses the number of MCMC sampling steps and is available only as a \emph{raw} benchmark.

\noindent\textbf{\paramnettime{}}~\citep{falkner-icml18a} This benchmark has $6$ architectural and training hyperparameters to train a feed-forward neural network on six different datasets from OpenML~\citep{vanschoren-sigkdd14a}: Adult, Higgs, Letter, MNIST, Optdigits and Poker. As fidelity it uses the number of training epochs for the neural networks. This is a surrogate benchmark and uses a random forest, which is trained on $10$K randomly samples configurations.

\noindent\textbf{\NASHPO{}.}~\citep{klein-arxiv19a} This benchmark is a joint neural architecture search and HPO for a 2-layer feedforward neural network. The output layer was designed as a linear layer with parameterized architecture details and training parameters while the search space is a large grid of configurations on four popular UCI datasets for regression: protein structure, slice localization, naval propulsion and parkinsons telemonitoring. 

\noindent\textbf{\NASOOO{}.}~\citep{ying-icml19a} This was the first introduced NAS benchmark based on tabular lookup, designed for reproducibility in NAS research. Each architecture is represented as a stack of architectural cells, where each such cell is represented as directed acyclic graphs (DAGs). The benchmarks offers a search space that includes nearly 423k unique architectures by parameterizing the nodes and edges of the DAGs. The lookup table allows to query performance of architectures on the Cifar-10 dataset. Additionally, queries can be made for intermediate training epochs too, thereby allowing \mf{} optimization. In contrast to the original implementation, we always return the average across the three repetitions as a score.

\noindent\textbf{\NASOSO{}.}~\citep{zela-iclr20b} The NAS-Bench-1shot1 was derived from the large architecture space of NAS-Bench-101, such that, weight-sharing based one-shot NAS methods can be applied for this tabular lookup. The cell-level encoding was modified to yield 3 variants of the architecture space which contains around 6k (search space 1), 29k (search space 2), 300k (search space 3) architectures. In contrast to the original implementation we always return the average across the three repetitions as a score.

\noindent\textbf{\NASTOO{}.}~\citep{dong-iclr20a} To further aid the use of weight sharing algorithms to NAS Benchmarks, this benchmark introduced a fixed cell search space wherein a DAG has only 4 nodes that define the cell architecture. Whereas the edges define the operations. Thus, creating a search space of around 15k unique architectures. NAS-Bench-201 provides a lookup table for Cifar-10, Cifar-100, and ImageNet16-120. In contrast to the original implementation we always return the average across the three repetitions as a score.

\subsection{New Benchmarks}
\label{app:ssec:new}

Here, we provide more details on the \emph{new} benchmarks and list their hyperparameter and fidelity spaces in Table~\ref{app:tab:space:new}.

\noindent\textbf{\svm{}} A $2$-dimensional benchmark for a SVM model with an RBF kernel with the \emph{regularization} and the kernel coefficient \emph{gamma} as available hyperparameters to tune. It uses the dataset subset fraction as the fidelity and is available as both \emph{raw} and \emph{tabular} benchmarks. For the tabular version, we discretized each hyperparameter into $21$ bins for $441$ unique hyperparameter configurations and evaluated each of these on $20$ datasets from the \automl{} benchmark~\citep{gijsbers-automl19a}.

\noindent\textbf{\lr{}} This benchmark has $2$ hyperparameters -- learning rate and regularization for a logistic regression model trained using Stochastic Gradient Descent (SGD). It uses dataset fraction and/or the number of SGD iterations as the fidelity and is available as both a \emph{raw} and \emph{tabular} benchmark. For the tabular version we evaluated a grid of $625$ configurations on $20$ datasets from the \automl{} benchmark~\citep{gijsbers-automl19a}.

\noindent\textbf{\xgb{}} This benchmark has $4$ hyperparameters that tune the maximum depth per tree, the features subsampled per tree, the learning rate and the L2 regularization for the XGBoost model. It uses dataset fraction and/or the number of boosting iterations as fidelities and is available as both a \emph{raw} and \emph{tabular} benchmark. For the tabular version we discretized each hyperparameter into $10$ bins and evaluated the resulting grid of $10k$ configurations on $20$ datasets from the \automl{} benchmark~\citep{gijsbers-automl19a}.

\noindent\textbf{\rf{}} This benchmark has $4$ hyperparameters that tune the maximum depth per tree, the maximum features subsampled per split, the minimum number of samples required for splitting a node, and the minimum number of samples required in each leaf node for a random forest model. It uses dataset fraction and/or the number of trees as fidelities and is available as both a \emph{raw} and \emph{tabular} benchmark. For the tabular version we discretized each hyperparameter into $10$ bins and evaluated the resulting grid of $10k$ configurations on $20$ datasets from the \automl{} benchmark~\citep{gijsbers-automl19a}.

\noindent\textbf{\mlp{}} This benchmark has $5$ hyperparameters -- two hyperparameters that determine the depth and width of the network; three more hyperparameters tune the batch size, L2 regularization and the initial learning rate for Adam. It uses dataset fraction and/or the number of epochs as fidelities and is available as both a \emph{raw} and \emph{tabular} benchmark. For the tabular version, we discretized each hyperparameter into $10$ bins and evaluated the resulting grid of $1k$ configurations for each of $30$ different architectures, resulting in $30k$ configurations in total, on $8$ datasets from the \automl{} benchmark~\citep{gijsbers-automl19a}.

To collect the data for the tabular benchmark, we evaluated every configuration-fidelity pair in the discretized space on $5$ different seeds; each such repetition is evaluated on the following $4$ metrics: \textit{accuracy}, \textit{balanced accuracy}, \textit{precision}, \textit{f1}.

\begin{minipage}[t]{0.63\textwidth}
    \centering
    \small
   \captionof{table}{Table detailing the configuration spaces for the \emph{new} benchmarks included in \hpobench{}. For each model, we report the hyperparameters and their ranges (top part) and fidelities and their ranges (bottom part).}
    \label{app:tab:space:new}
    \begin{tabular}{@{\hskip 0mm}l@{\hskip 1mm}l@{\hskip 1mm}l@{\hskip 2mm}l@{\hskip 1mm}r@{\hskip 0mm}}
    \toprule
    benchmark &  name & type & log & range \\
    \midrule
    \multirow{3}{*}{\svm{}} & C & float & \yesmark & $[2^{-10}, 2^{10}]$ \\
    & gamma & float & \yesmark & $[2^{-10}, 2^{10}]$ \\
    \cmidrule{2-5}
     &  \textit{subsample} &   float &  \nomark & $[0.1, 1.0]$ \\
    \midrule
     \multirow{4}{*}{\lr{}} & alpha & float & \yesmark & $[1e-05, 1.0]$ \\
      & eta0 & float & \yesmark & $[1e-05, 1.0]$ \\
    \cmidrule{2-5}
     & iter &   int &  \nomark & $[10, 1000]$ \\
     & subsample &   float &  \nomark & $[0.1, 1.0]$ \\
     \midrule
      \multirow{6}{*}{\xgb{}} & colsample\_bytree & float & \nomark & [0.1, 1.0] \\
       & eta & float &  \yesmark & $[2^{-10}, 1.0]$ \\
       & max\_depth &   int &  \yesmark & $[1, 50]$ \\
      & reg\_lambda & float &  \yesmark & $[2^{-10}, 2^{10}]$ \\
    \cmidrule{2-5}
     & n\_estimators &   int &  \nomark & $[50, 2000]$ \\
     & subsample &   float &  \nomark & $[0.1, 1.0]$ \\
    \midrule
     \multirow{6}{*}{\rf{}} & max\_depth &   int &  \yesmark & $[1, 50]$ \\
      & max\_features & float & \nomark & $[0.0, 1.0]$ \\
      & min\_samples\_leaf &   int & \nomark & $[1, 2]$ \\
      & min\_samples\_split &   int &  \yesmark & $[2, 128]$ \\
    \cmidrule{2-5}
     & n\_estimators &   int &  \nomark & $[16, 512]$ \\
     & subsample &   float &  \nomark & $[0.1, 1.0]$ \\
    \midrule
    \multirow{7}{*}{\mlp{}} & alpha & float &  \yesmark & $[1.0e^{-08}, 1.0]$ \\
      & batch\_size &   int &  \yesmark & $[4, 256]$ \\
      & depth &   int & \nomark & $[1, 3]$ \\
      & learning\_rate\_init & float &  \yesmark & $[1.0e^{-05}, 1.0]$ \\
      & width &   int &  \yesmark & $[16, 1024]$ \\
     \cmidrule{2-5}
     & epochs &   int &  \nomark & $[3, 243]$ \\
     & subsample &   float &  \nomark & $[0.1, 1]$ \\
    \bottomrule
\end{tabular}
\end{minipage}%
\hfill
\begin{minipage}[t]{0.33\textwidth}
\small
\centering
\captionof{table}{OpenML Task IDs used from the \automl{} benchmark for \svm{}, \lr{}, \xgb{} and \rf{}. \mlp{} uses only the first $8$ task IDs. The table shows the total number of instances available (train + test) (\#obs), and the total number of features prior to preprocessing (\#feat).} 
\label{tab:tids}
\begin{tabular}{@{\hskip 0mm}l@{\hskip 1mm}r@{\hskip 1mm}r@{\hskip 1mm}r@{\hskip 0mm}}
\toprule
name &     tid &   \#obs &  \#feat \\
\midrule
\href{https://www.openml.org/t/10101}{blood-transf..} &   10101 &    748 &      4 \\
\href{https://www.openml.org/t/53}{vehicle} &      53 &    846 &     18 \\
\href{https://www.openml.org/t/146818}{Australian} &  146818 &    690 &     14 \\
\href{https://www.openml.org/t/146821}{car} &  146821 &   1728 &      6 \\
\href{https://www.openml.org/t/9952}{phoneme} &    9952 &   5404 &      5 \\
\href{https://www.openml.org/t/146822}{segment} &  146822 &   2310 &     19 \\
\href{https://www.openml.org/t/31}{credit-g} &      31 &   1000 &     20 \\
\href{https://www.openml.org/t/3917}{kc1} &    3917 &   2109 &     22 \\
\midrule
\href{https://www.openml.org/t/168912}{sylvine} &  168912 &   5124 &     20 \\
\href{https://www.openml.org/t/3}{kr-vs-kp} &       3 &   3196 &     36 \\
\href{https://www.openml.org/t/167119}{jungle\_che..} &  167119 &  44819 &      6 \\
\href{https://www.openml.org/t/12}{mfeat-factors} &      12 &   2000 &    216 \\
\href{https://www.openml.org/t/146212}{shuttle} &  146212 &  58000 &     9 \\
\href{https://www.openml.org/t/168911}{jasmine} &  168911 &   2984 &    145 \\
\href{https://www.openml.org/t/9981}{cnae-9} &    9981 &   1080 &    856 \\
\href{https://www.openml.org/t/167120}{numerai28.6} &  167120 &  96320 &     21 \\
\href{https://www.openml.org/t/14965}{bank-mark..} &   14965 &  45211 &     16 \\
\href{https://www.openml.org/t/146606}{higgs} &  146606 &  98050 &     28 \\
\href{https://www.openml.org/t/7592}{adult} &    7592 &  48842 &     14 \\
\href{https://www.openml.org/t/9977}{nomao} &    9977 &  34465 &    118 \\
\bottomrule
\end{tabular}
\end{minipage}%

\section{Details on Hardware Used for Experiments}
\label{app:hardware}
For our benchmark study we ran all jobs on a compute cluster equipped with Intel(R) Xeon(R) Gold 6242 CPU @ 2.80GHz. If not stated otherwise, we run all job on $1$ CPU with up to $6$GB RAM for at most $4$ days or till the benchmark budget was exhausted. For runs that needed more memory to load data, we allowed up to $12$GB RAM (\NASOOO{}, \NASOSO{}, \NASTOO{}). For collecting tabular data for the \emph{new} benchmarks, we ran all jobs on a compute cluster equipped with Intel(R) Broadwell E5-2630v4 @ 2.2GHz with up to $6$GB RAM.

\section{Details on Runtime}
\label{app:runtime}

Running all optimizers on the raw versions of the existing community benchmarks would take more than $1500$ CPU years, but the use of tabular and surrogate-based benchmarks in \hpobench{} reduces this amount to \emph{only} $22.5$ CPU years. While this is still a lot, we emphasize that most of this time is used by the optimizers (and not the benchmarks). For developing and evaluating a new \mf{} method and comparing it to computationally cheap baselines, e.g. sequentially evaluating both \random{} and \de{} on all tabular and surrogate benchmarks took $<10$ CPU days, \hb{} took around $50$ CPU days and \dehb{} needed around $40$ CPU days.
To further explain the amount of time it took to obtain results for our empirical study, we look at statistics of our runs. In Table~\ref{tab:app:timestats}, we report the average runtime (in hours, maximum $96$, however, we only record the last call to our objective function, so a runtime of, e.g. $95$ could also mean that the optimizer did not call the objective function for $2$ hours and was then forcefully terminated) and the number of calls$/100$ to the objective function for one exemplary benchmark per family. The last two rows show the total time spent on obtaining results for all raw benchmarks and surrogate plus tabular benchmarks per optimizers. Additionally, we give the overall amount of compute spent on our empirical study.

Looking at the first part of the table, we favourably see, that most optimizers on average took less than two hours to spend the simulated optimization budget. However, there are some exceptions like \smacbo{} and \dragonfly{} mostly hitting the optimization budget of $4$ days resulting in fewer calls to the objective function and worse performance.

Additionally, these statistics also allow to study some failure cases of the optimizers. For \dragonfly{} on \NASOSO{}, it only evaluated $90$ configurations while taking less than $1$ hour. Here \dragonfly{} stopped right after the initial design, because it could not construct a model, the same happened for the \pybnn{} benchmarks and thus the total runtime for the raw benchmarks is substantially lower. Finally, \random{} called \NASA{} three times more often than other \bb{} optimizers, because the table underlying this benchmark does not cover the complete hyperparameter space and thus returns a loss of $1$ and costs of $0$ for configurations not in the table. More advanced search algorithms avoid these seemingly badly performing regions and thus sample more \emph{costly} evaluations.

\begin{table}[htbp]
    \small
    \centering\small
    \caption{We report the median wallclock time (in hours) and number of calls$/100$ to the objective function for all optimizers and one benchmark per benchmark family.}
    \label{tab:app:timestats}
\begin{tabular}{@{\hskip 0mm}l@{\hskip 2mm}
cc@{\hskip 2mm}
cc@{\hskip 2mm}
cc@{\hskip 2mm}
cc@{\hskip 2mm}
cc@{\hskip 2mm}
cc@{\hskip 0mm}}
\toprule
& \multicolumn{2}{c}{\paramadult{}} & \multicolumn{2}{c}{\slice{}} & \multicolumn{2}{c}{\NASA{}} & \multicolumn{2}{c}{\nbcifarh{}} & \multicolumn{2}{c}{\NASOSOA{}} & \multicolumn{2}{c}{total time}\\
 optimizer & t & \#c & t & \#c & t & \#c & t & \#c & t & \#c & raw & tab+sur \\
\midrule
\random{} & 0 & 25 & 0 & 47 & 0 & 118 & 0 & 9 & 0 & 23 & 2295 & 102 \\
\de{} & 0 & 25 & 0 & 29 & 0 & 31 & 0 & 7 & 0 & 13 & 2290 & 48 \\
\tpe{} & 0 & 25 & 1 & 31 & 0 & 31 & 0 & 8 & 0 & 27 & 2297 & 716 \\
\smacbo{} & 82 & 13 & 96 & 12 & 96 & 7 & 9 & 7 & 96 & 12 & 2296 & 46566 \\
\smacrf{} & 2 & 25 & 2 & 43 & 2 & 39 & 0 & 8 & 1 & 21 & 2299 & 7326 \\
\hebo{} & 84 & 13 & 96 & 12 & 96 & 11 & 26 & 7 & 96 & 12 & 2300 & 48735 \\
\midrule
\hb{} & 0 & 108 & 3 & 209 & 1 & 242 & 0 & 21 & 0 & 95 & 2300 & 1299 \\
\bohb{} & 0 & 108 & 2 & 130 & 0 & 104 & 0 & 20 & 1 & 110 & 2298 & 1508 \\
\dehb{} & 0 & 108 & 0 & 126 & 0 & 118 & 0 & 17 & 0 & 63 & 2255 & 1031 \\
\smachb{} & 8 & 105 & 8 & 202 & 8 & 166 & 0 & 20 & 2 & 96 & 2298 & 12370 \\
\dragonfly{} & 93 & 10 & 92 & 9 & 90 & 6 & 94 & 10 & 0 & 1 & 532 & 41867 \\
\optunath{} & 0 & 109 & 0 & 109 & 0 & 75 & 0 & 31 & 0 & 55 & 2297 & 1606 \\
\optunatm{} & 4 & 287 & 0 & 111 & 0 & 81 & 0 & 15 & 0 & 78 & 2278 & 5694 \\
\midrule
\multicolumn{11}{r}{sum in CPU years} & 3.2 & 19.3 \\
\bottomrule
\end{tabular}
\end{table}

\section{More Details on Considered Optimizers}
\label{app:opt}

Here, we provide additional details on the optimizers used in this work. We provide an overview in Table~\ref{tab:app:optimizers} and then briefly explain our baselines, \bb{} and \mf{} optimizers in detail. We note that we used the default settings for all tools and implementations.

\begin{table}[htbp]
    \centering
    \small
    \caption{Overview of HPO optimizers considered in this study. For each optimizer we list the model type, what types of hyperparameters and fidelities it can handle (the tool can either handle it natively (\yesmark), not handle it (\nomark) or we could transform the type (\maybemark)), a link to the codebase and references.}
    \label{tab:app:optimizers}
    \begin{tabular}{lccccccccc}
\toprule
        \multirow{2}{*}{name} & \multirow{2}{*}{model} & \multicolumn{3}{c}{types} & \multicolumn{2}{c}{fidelities} & \multirow{2}{*}{link} & \multirow{2}{*}{reference} & \multirow{2}{*}{version} \\
             &       & cont & cat & log & disc. & cont. &  &  \\
\midrule
    \random{} & - & \yesmark & \yesmark & \yesmark & \nomark & \nomark & - & \citep{bergstra-jmlr12a} \\
    \smacbo{} & GP & \yesmark & \yesmark & \yesmark & \nomark & \nomark & \href{https://github.com/automl/SMAC3}{SMAC3} & \cite{hutter-lion11a,lindauer-jmlr22a} & 1.0.1 \\
    \smacrf{} & RF & \yesmark & \yesmark & \yesmark & \nomark & \nomark & \href{https://github.com/automl/SMAC3}{SMAC3} & \cite{hutter-lion11a,lindauer-jmlr22a} & 1.0.1 \\
    \tpe{} & KDE & \yesmark & \yesmark & \yesmark & \nomark & \nomark & \href{https://github.com/automl/HpBandSter}{HpBandSter} & \cite{falkner-icml18a} & 0.7.4 \\
    \de{} & - & \yesmark & \yesmark & \yesmark & \nomark & \nomark & \href{https://github.com/automl/DEHB}{DEHB} & \cite{awad-iclr20} & \href{https://github.com/automl/DEHB/commit/2e4ee7754d90d7196f1148e4b1c252db60be58fa}{git commit} \\
    \hebo{} & GP & \yesmark & \yesmark & \yesmark & \nomark & \nomark &  \href{https://github.com/huawei-noah/HEBO/tree/master/HEBO}{HEBO} & \cite{cowenrivers-jair22a} & 0.1.0 \\
\midrule
    \hb{} & - & \yesmark & \yesmark & \yesmark & \maybemark & \yesmark & 
    \href{https://github.com/automl/HpBandSter}{HpBandSter} & \cite{li-jmlr18a} & 0.7.4 \\
    \bohb{} & KDE & \yesmark & \yesmark & \yesmark & \maybemark & \yesmark & \href{https://github.com/automl/HpBandSter}{HpBandSter} & \cite{falkner-icml18a} & 0.7.4 \\
    \dehb{} & - & \yesmark & \yesmark & \yesmark & \maybemark & \yesmark & \href{https://github.com/automl/DEHB}{DEHB} & \cite{awad-ijcai21} & \href{https://github.com/automl/DEHB/commit/2e4ee7754d90d7196f1148e4b1c252db60be58fa}{git commit} \\
    \smachb{} & RF & \yesmark & \yesmark & \yesmark & \maybemark & \yesmark & 
    \href{https://github.com/automl/SMAC3}{SMAC3} & \cite{hutter-lion11a,lindauer-jmlr22a} & 1.0.1 \\
    \dragonfly{} & GP & \yesmark & \yesmark & \maybemark & \yesmark & \yesmark &  \href{https://github.com/dragonfly/dragonfly}{Dragonfly} & \cite{kandasamy-jmlr20a} & 0.1.5 \\
    \optunatm{} & TPE & \yesmark & \yesmark & \yesmark & \yesmark & \nomark & \href{https://optuna.readthedocs.io/en/stable/}{Optuna} & \cite{akiba-kdd19a} & 2.8.0 \\
    \optunath{} & TPE & \yesmark & \yesmark & \yesmark & \yesmark & \nomark & \href{https://optuna.readthedocs.io/en/stable/}{Optuna} & \cite{akiba-kdd19a} & 2.8.0 \\
\bottomrule
\end{tabular}
\end{table}

\subsection{Baselines}
\label{ssec:exp:opt:base}

\noindent\textbf{Random Search (\random)} is a simple baseline that samples new configurations uniformly at random from a prior. It was proposed as an improved baseline over grid search~\cite{bergstra-jmlr12a} as it can handle low intrinsic dimensionality and is easier to run in parallel. 

\noindent\textbf{Hyperband (\hb{})}~\cite{li-jmlr18a} is a bandit algorithm for the pure-exploration, non-stochastic infinite-armed bandit problem which we described in Section~\ref{sec:background}. We will use it as a random search baseline for \mf{} optimization.

\subsection{Black-box Optimizers}
\label{ssec:exp:opt:black}

\noindent\textbf{\smacbo{}} is an implementation of traditional Gaussian process-based \bo{} with a Matérn kernel~\cite{snoek-nips12a} and a SOBOL sequence initial design~\cite{jones-jgo98a}. For categorical hyperparameters it uses a Hamming kernel~\citep{Hutter09} and is implemented in the SMAC toolbox~\citep{lindauer-jmlr22a}, thus it is using local search for acquisition function optimization~\citep{hutter-lion11a}. Its hyperparameters were tuned for good average performance over $50$ function evaluations using meta-optimization~\citep{lindauer-dso19a}.

\noindent\textbf{\smacrf{}} is similar to \smacbo{} but uses random forests as suggested in the original \smac{} publication~\cite{hutter-lion11a}. In contrast to the original hyperparameter setting of \smac{} with random forests, this version uses a SOBOL sequence initial design~\cite{jones-jgo98a} and only $20\%$ interleaved random samples instead of $50\%$. These hyperparameter settings were found via meta-optimization~\citep{lindauer-dso19a} for good average performance over 50 function evaluations.

\noindent\textbf{\tpe} is a re-implementation of the TPE algorithm using multi-dimensional kernel density estimators as used by the \bohb{} algorithm~\citep{falkner-icml18a}. Instead of modeling the objective function as $p(y|x)$, it models two densities, $p(x|y_{good})$ and $p(x|y_{bad})$, and uses their ratio that is proportional to the expected improvement acquisition function~\citep{bergstra-nips11a}.

\noindent\textbf{\de{}.} We use the canonical DE with \textit{rand/1} as the mutation strategy and \textit{binomial} crossover. We set the mutation factor $F$ and crossover rate $CR$ to $0.5$ each and the population size $NP$ to $20$~\citep{awad-iclr20}. 

\noindent\textbf{\hebo{}} is a GP-based BO algorithm that uses input warping and output warping, an ensemble of acquisition functions~\citep{cowenrivers-jair22a} and won the recent NeurIPS Blackbox Optimization challenge~\citep{turner-neuripscomp20a}.

\subsection{Multi-fidelity Optimizers}
\label{ssec:exp:opt:mf}

\noindent\textbf{\bohb{}}~\citep{falkner-icml18a} combines \bo{} and \hb{} with the goal of both algorithms complementing each other. It follows the regular \hb{} scheme, but instead of sampling configurations at random it uses \bo{}. For \bo{} it uses a KDE model as described above. To handle multiple fidelities it builds an independent model per fidelity, but only if there is sufficient (number of hyperparameters + $1$) training data available, to then always use the model from the highest fidelity for which a model is available.

\noindent\textbf{\smachb{}}~\citep{lindauer-jmlr22a} is a straight-forward re-implementation of the \bohb{} algorithm using the \smacrf{} building blocks described in the previous section.

\noindent\textbf{\dehb{}}~\citep{awad-ijcai21} is a new model-free successor of \bohb{} which uses the evolutionary optimization method DE instead of BO. For each fidelity, \dehb{} maintains a subpopulation and runs a separate DE evolution while the information about good configurations flows from subpopulations
at lower fidelities to those at higher fidelities through a
modified mutation strategy. The mutation allows the use of these good configurations from lower fidelities to be selected as parents to evolve the new subpopulation at a higher fidelity.
The hyperparameters of the \de{}-part of \dehb{} are set exactly as for \de{} described above.   

\noindent\textbf{Dragonfly (\dragonfly{})}~\citep{kandasamy-jmlr20a} is a BO algorithm which implements an improved version of the BOCA algorithm~\citep{kandasamy-icml17a}, which uses Gaussian processes and the upper confidence bound acquisition function to first decide a location to query before deciding the fidelity to query.

\noindent\textbf{\optunatm{}} is implemented in the Optuna framework~\citep{akiba-kdd19a}, which is a high level optimization framework that allows to combine sampling (to propose new configurations to evaluate) and pruning (to stop configurations if they are not promising) strategies to construct optimization algorithms. \optunatm{} uses TPE as a sampling algorithm and the median stopping~\citep{golovin-kdd17a} rule as a pruning algorithm. It fits a Gaussian Mixture Model on the best so far seen configurations. The pruner stops a configuration if its best intermediate result is worse compared to the median of the other configurations on the same fidelity level. 

\noindent\textbf{\optunath{}} is like \optunatm{} implemented in the Optuna framework~\citep{akiba-kdd19a} and uses TPE for sampling, but \hb{} as a pruning algorithm.

\section{More Results}
\label{app:moreresults}

\begin{figure}[htbp]
    \centering
    \begin{tabular}{@{\hskip 0mm}c@{\hskip 0mm}c@{\hskip 0mm}c@{\hskip 0mm}c@{\hskip 0mm}}
    \cartpole{} & \pybnn{} & \paramnettime{} & \NASHPO{} \\
 \includegraphics[width=0.24\textwidth]{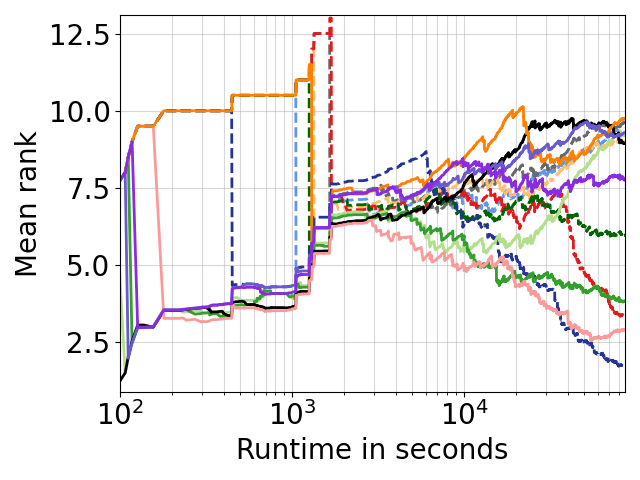} &
 \includegraphics[width=0.24\textwidth]{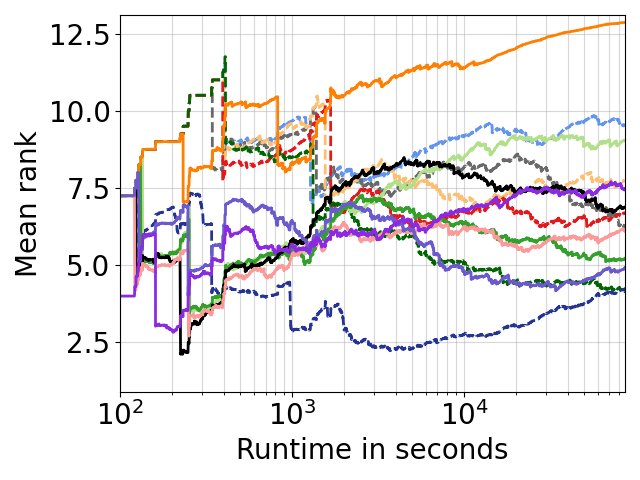} &
  \includegraphics[width=0.24\textwidth]{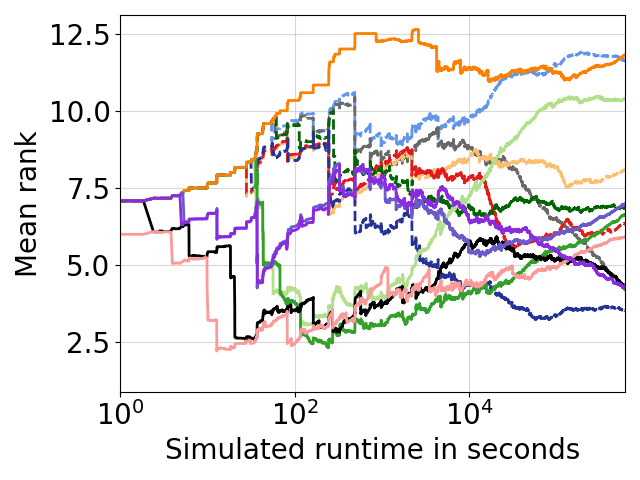} &
 \includegraphics[width=0.24\textwidth]{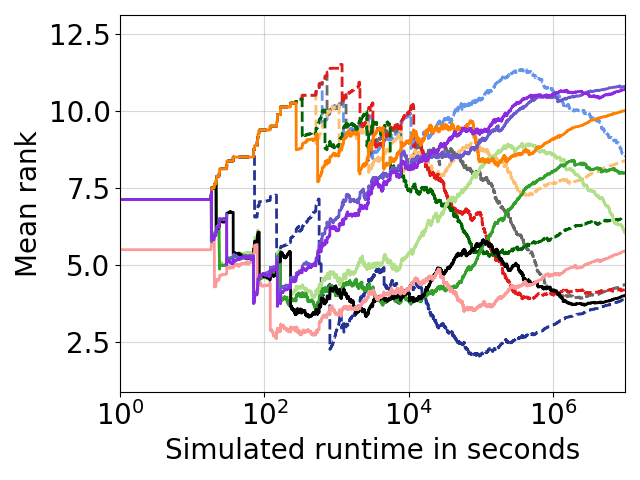} \\
 \NASOOO & \NASOSO & \NASTOO & \svm{} \\
  \includegraphics[width=0.24\textwidth]{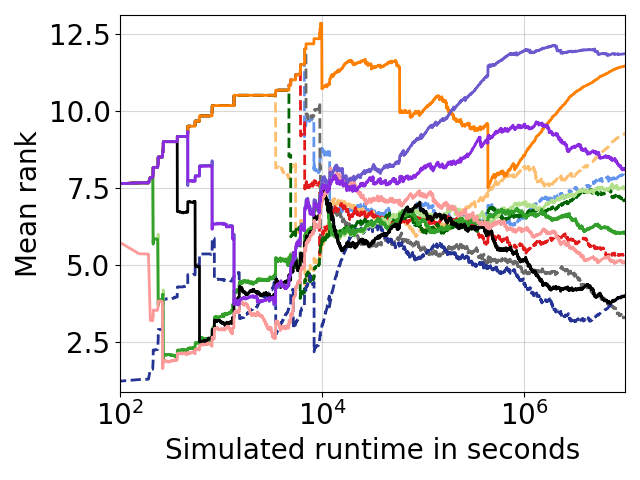} &
 \includegraphics[width=0.24\textwidth]{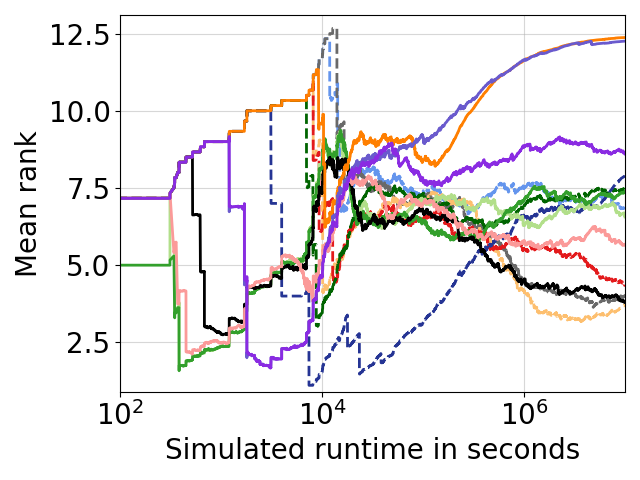} &
 \includegraphics[width=0.24\textwidth]{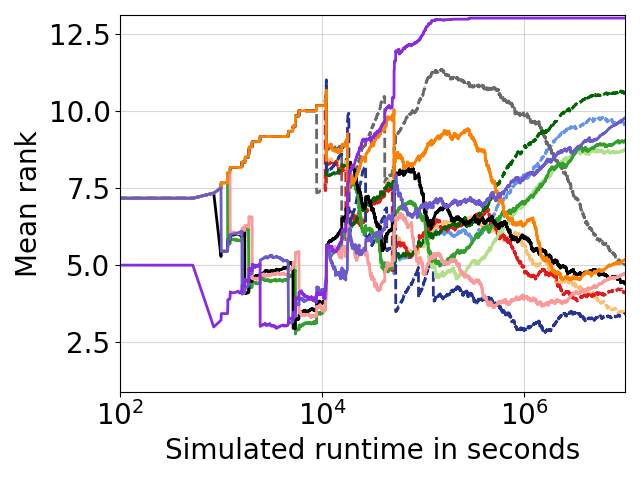} &
 \includegraphics[width=0.24\textwidth]{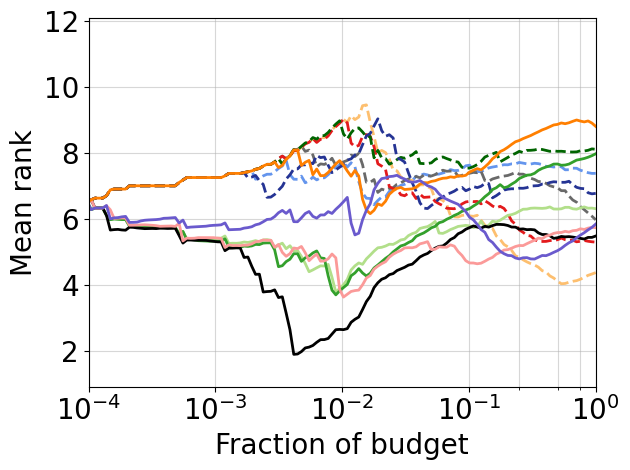} \\ 
 \lr{} & \rf{} & \xgb{} & \mlp{} \\
   \includegraphics[width=0.24\textwidth]{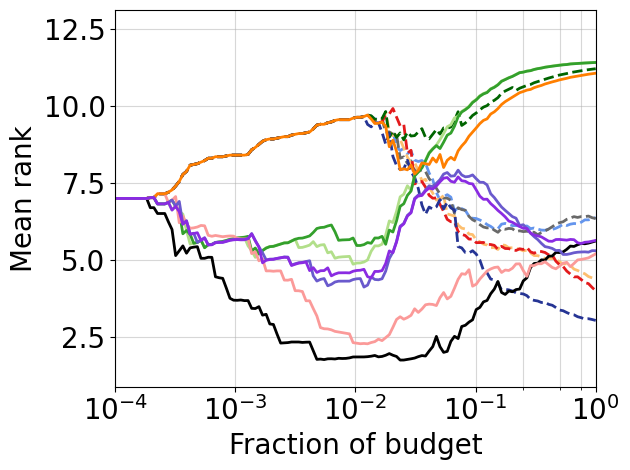} &
 \includegraphics[width=0.24\textwidth]{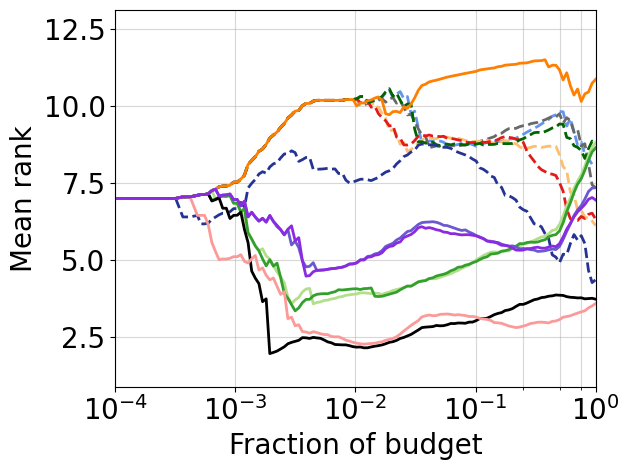} &
 \includegraphics[width=0.24\textwidth]{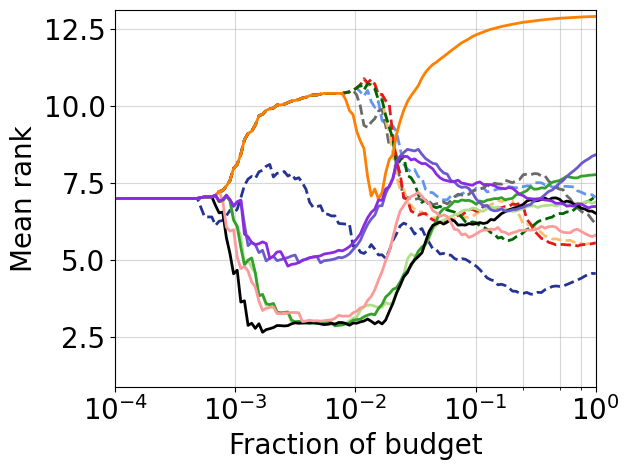} &
 \includegraphics[width=0.24\textwidth]{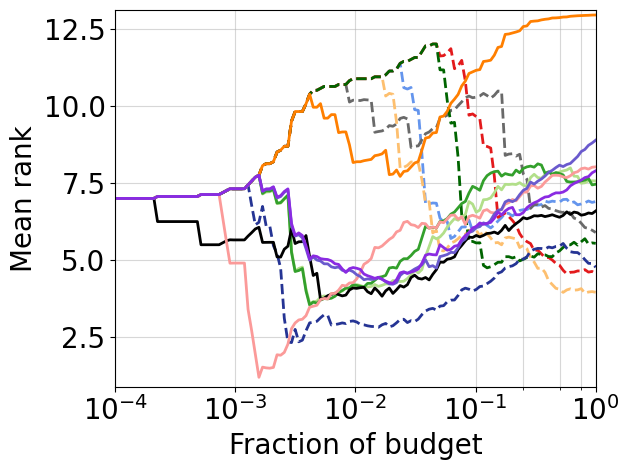} \\
 \multicolumn{4}{c}{
 \includegraphics[width=0.9\textwidth]{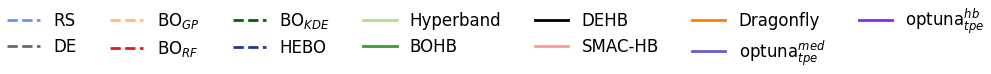} 
 } \\
    \end{tabular}
    \caption{Median rank over time. We report the median rank of the performance across all benchmarks of a benchmark family (see Table~\ref{tab:benchmarks}) for \textbf{all} optimizers.
    \label{fig:app:rankall}}
\end{figure}

\begin{figure}[htbp]
    \centering
    \begin{tabular}{@{\hskip 0mm}c@{\hskip 0mm}c@{\hskip 0mm}c@{\hskip 0mm}c@{\hskip 0mm}}
    \cartpole{} & \pybnn{} & \paramnettime{} & \NASHPO{} \\
 \includegraphics[width=0.24\textwidth]{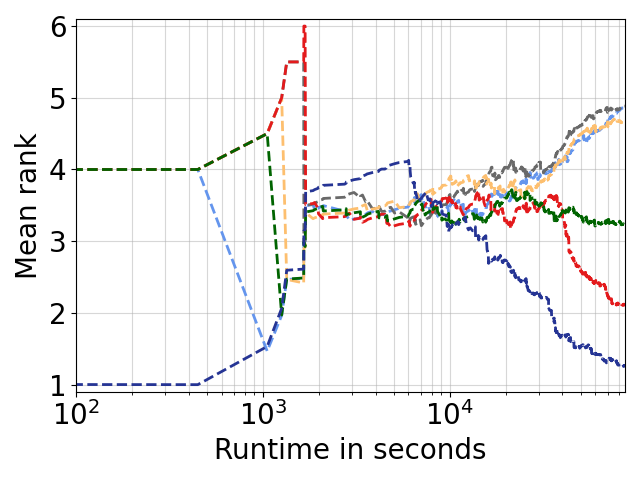} &
 \includegraphics[width=0.24\textwidth]{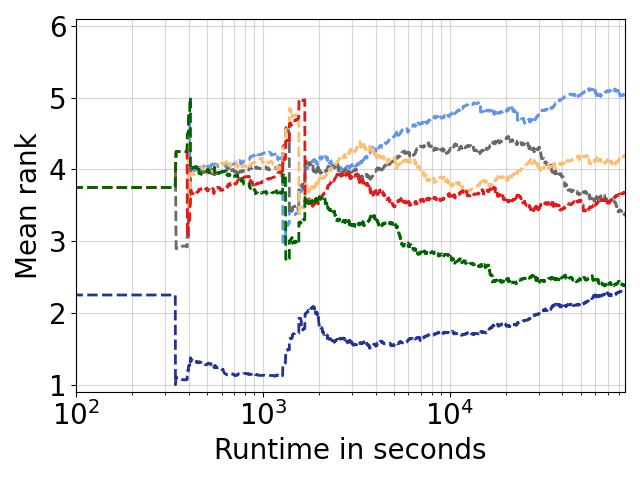} &
  \includegraphics[width=0.24\textwidth]{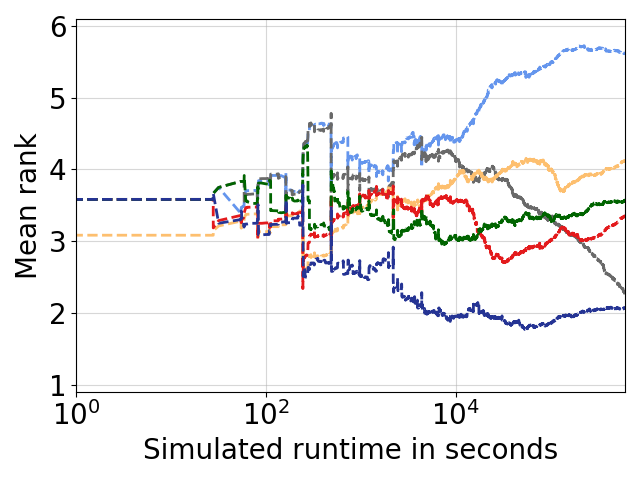} &
 \includegraphics[width=0.24\textwidth]{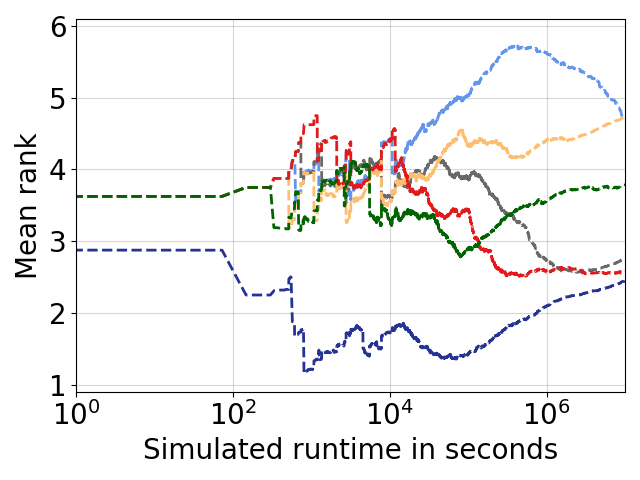} \\
 \NASOOO & \NASOSO & \NASTOO & \svm{} \\
  \includegraphics[width=0.24\textwidth]{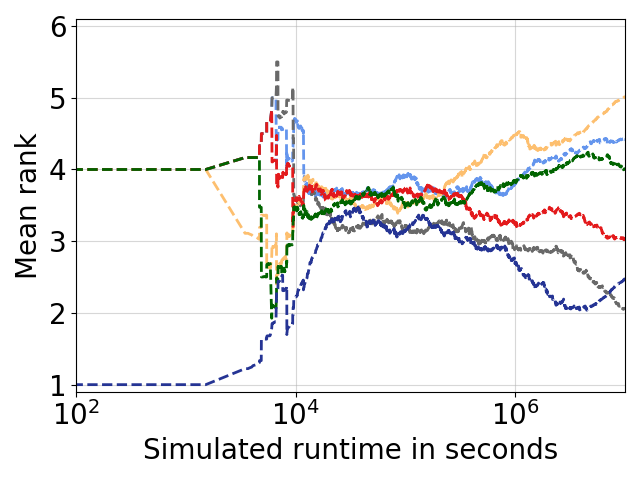} &
 \includegraphics[width=0.24\textwidth]{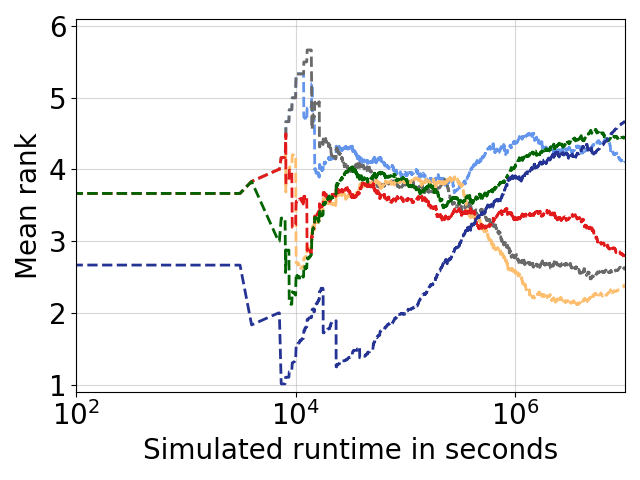} &
 \includegraphics[width=0.24\textwidth]{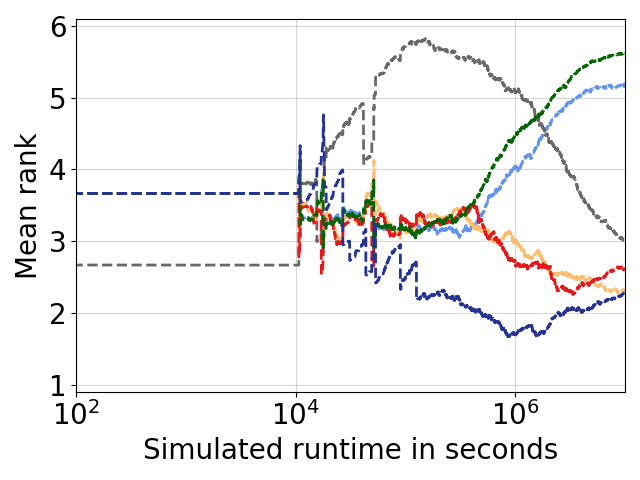} &
 \includegraphics[width=0.24\textwidth]{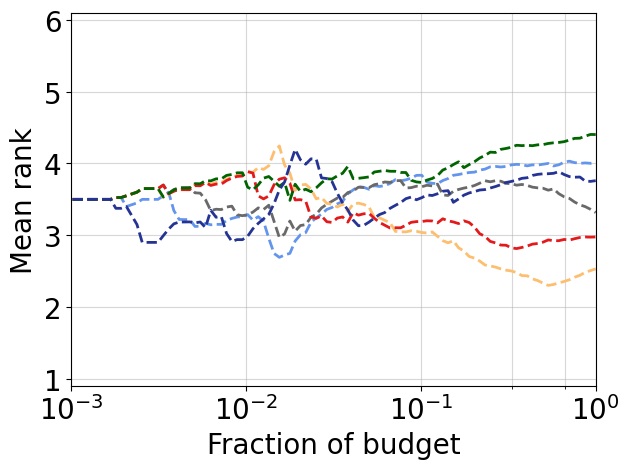} \\
 \lr{} & \rf{} & \xgb{} & \mlp{} \\
\includegraphics[width=0.24\textwidth]{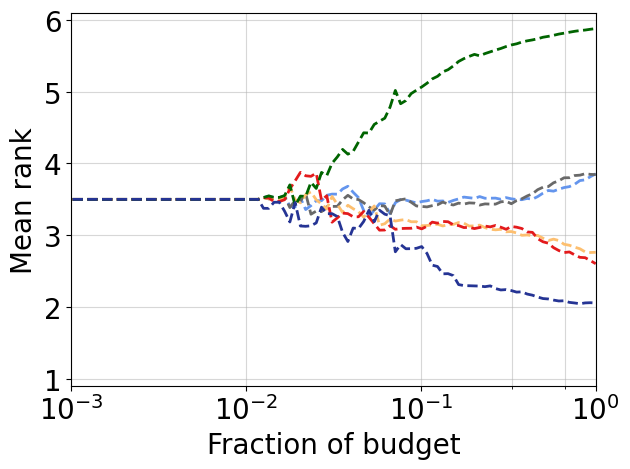} &
 \includegraphics[width=0.24\textwidth]{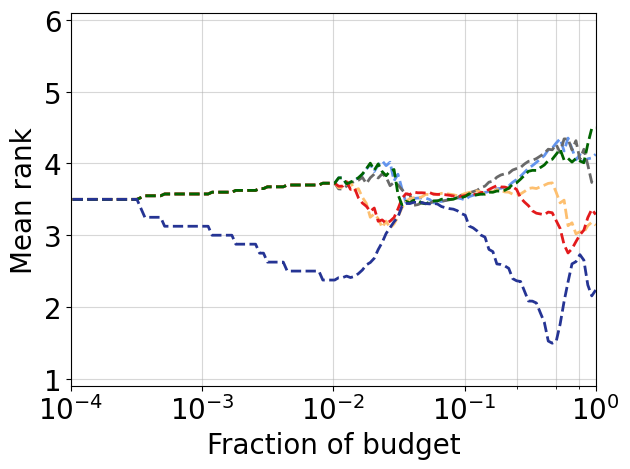} &
 \includegraphics[width=0.24\textwidth]{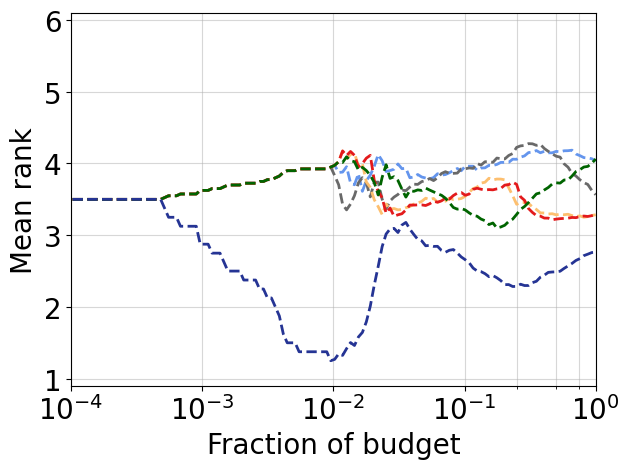} &
 \includegraphics[width=0.24\textwidth]{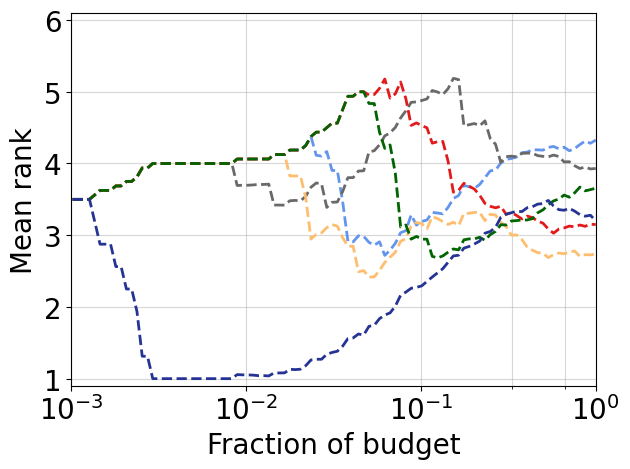} \\
 \multicolumn{4}{c}{
 \includegraphics[width=0.9\textwidth]{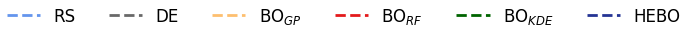} 
 } \\
    \end{tabular}
    \caption{Median rank over time. We report the median rank of the performance across all benchmarks of a benchmark family (see Table~\ref{tab:benchmarks}) for \textbf{\bb{}} optimizers.
    \label{fig:app:ranksf}}
\end{figure}

\begin{figure}[htbp]
    \centering
    \begin{tabular}{@{\hskip 0mm}c@{\hskip 0mm}c@{\hskip 0mm}c@{\hskip 0mm}c@{\hskip 0mm}}
    \cartpole{} & \pybnn{} & \paramnettime{} & \NASHPO{} \\
 \includegraphics[width=0.24\textwidth]{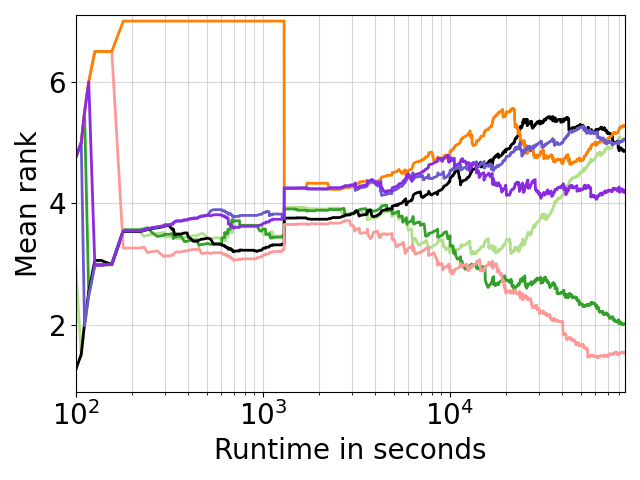} &
 \includegraphics[width=0.24\textwidth]{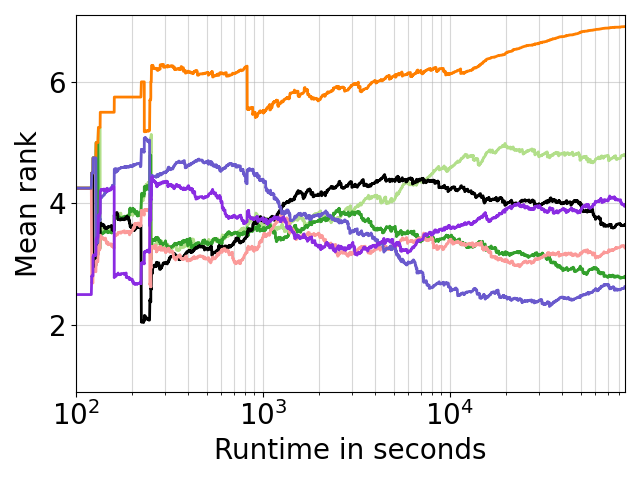} &
  \includegraphics[width=0.24\textwidth]{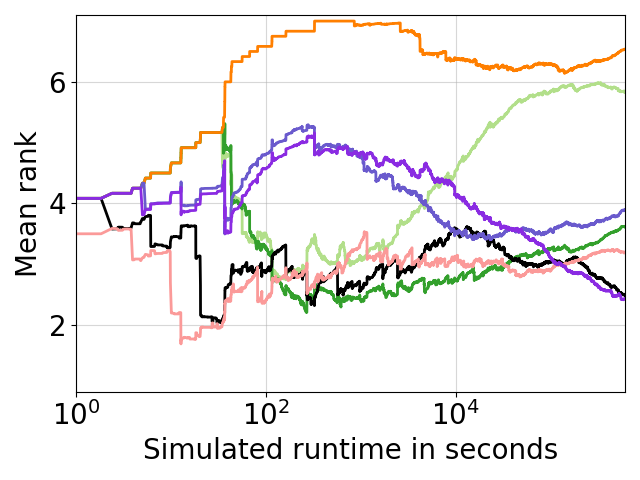} &
 \includegraphics[width=0.24\textwidth]{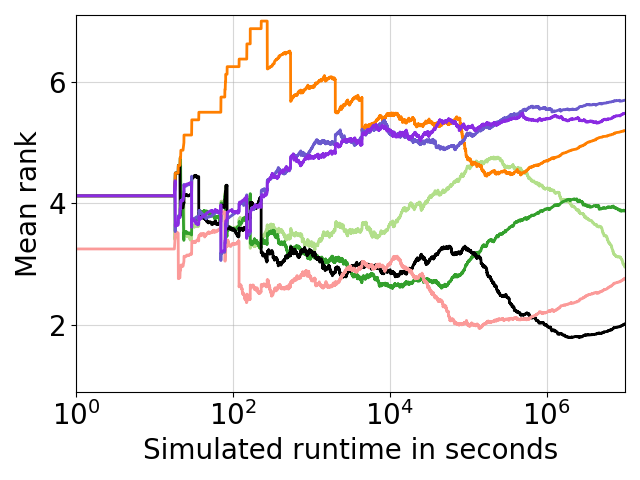} \\
 \NASOOO & \NASOSO & \NASTOO  & \svm{} \\
  \includegraphics[width=0.24\textwidth]{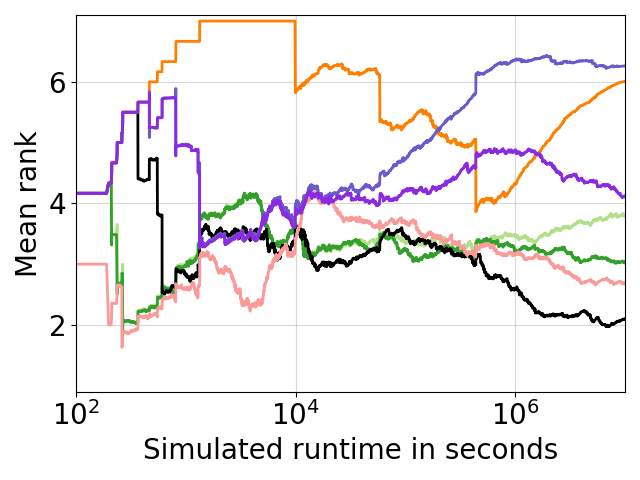} &
 \includegraphics[width=0.24\textwidth]{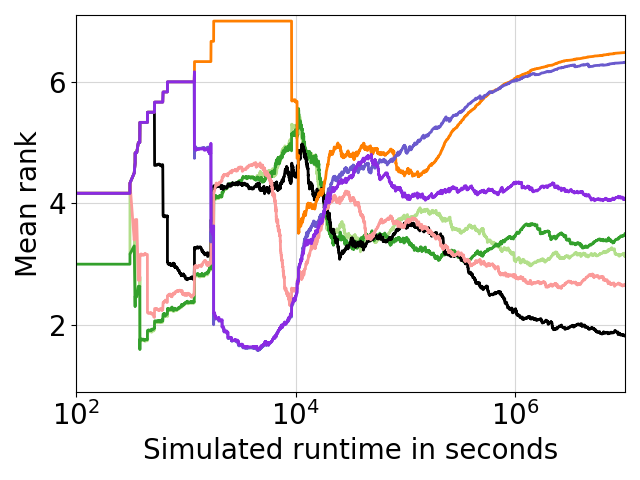} &
 \includegraphics[width=0.24\textwidth]{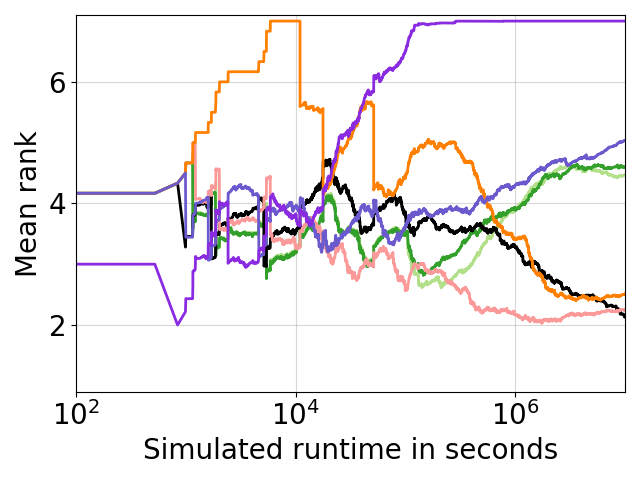} &
 \includegraphics[width=0.24\textwidth]{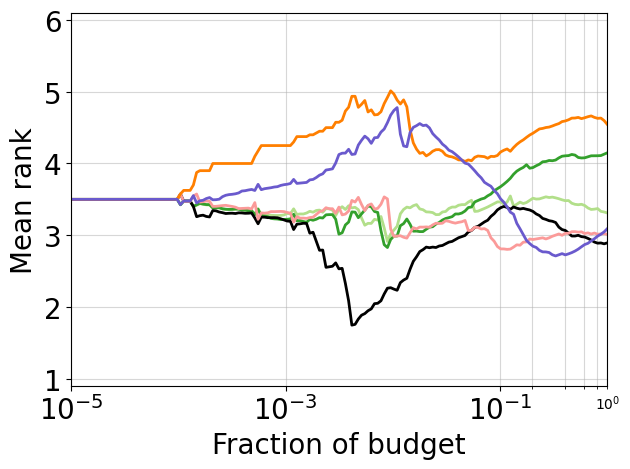} \\ 
 \lr{} & \rf{} & \xgb{} & \mlp{} \\
   \includegraphics[width=0.24\textwidth]{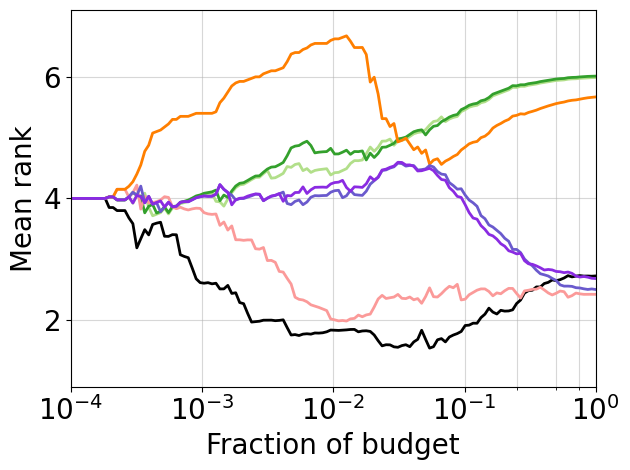} &
 \includegraphics[width=0.24\textwidth]{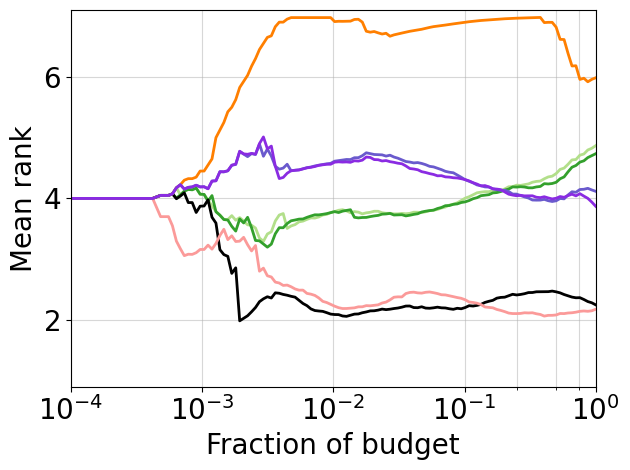} &
 \includegraphics[width=0.24\textwidth]{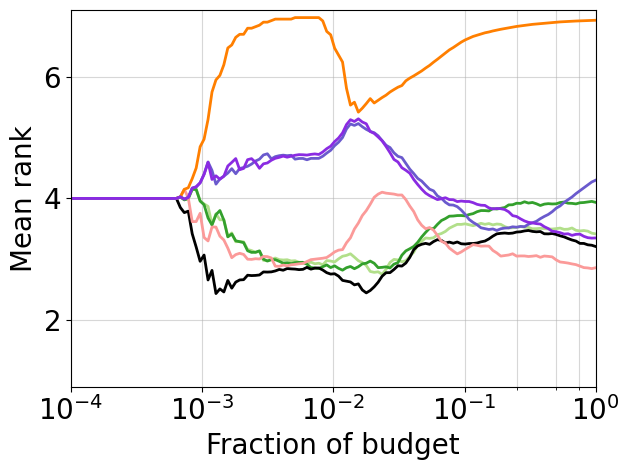} &
 \includegraphics[width=0.24\textwidth]{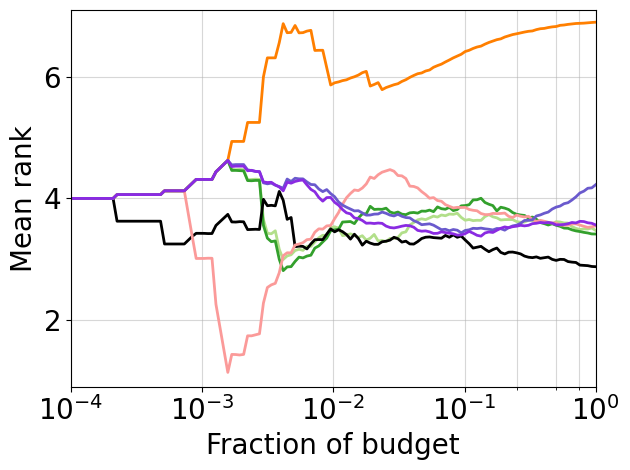} \\
 \multicolumn{4}{c}{
 \includegraphics[width=0.9\textwidth]{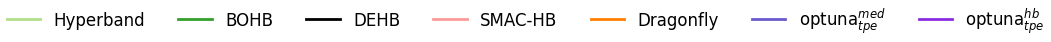} 
 } \\
    \end{tabular}
    \caption{Median rank over time. We report the median rank of the performance across all benchmarks of a benchmark family (see Table~\ref{tab:benchmarks}) for \textbf{\mf{}} optimizers.
    \label{fig:app:rankmf}}
\end{figure}

\begin{figure}[p]
    \centering
    \begin{tabular}{@{\hskip 0mm}c@{\hskip 0mm}c@{\hskip 0mm}c@{\hskip 0mm}c@{\hskip 0mm}}
    \cartpole & \bnnprotein & \bnnyear \\
    \includegraphics[width=0.24\textwidth]{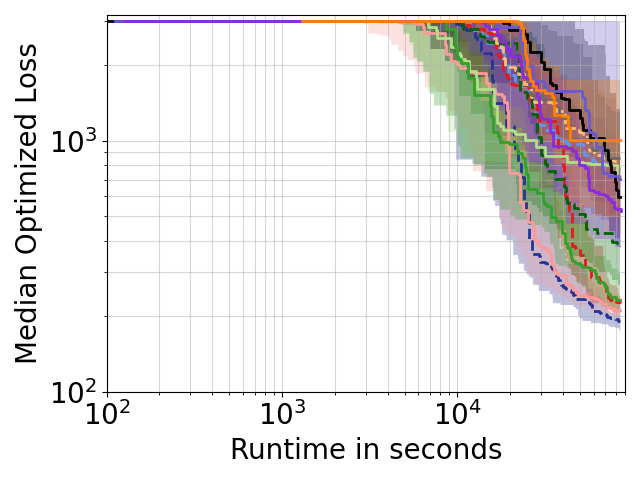} &
    \includegraphics[width=0.24\textwidth]{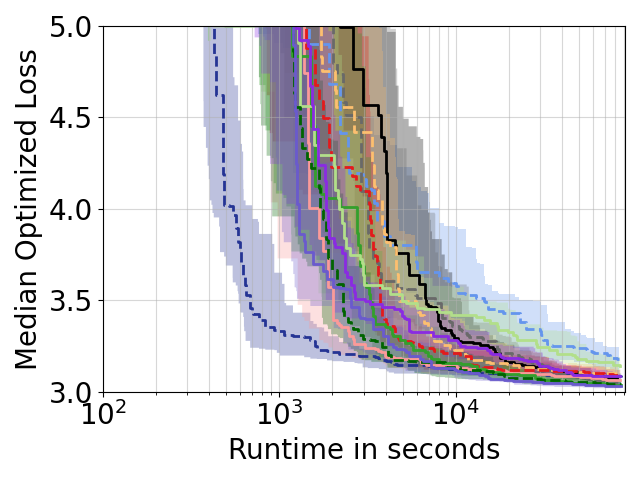} &
    \includegraphics[width=0.24\textwidth]{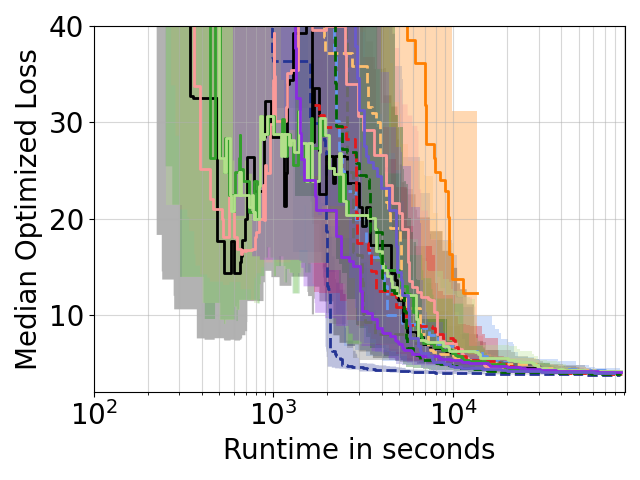} \\
    \paramadult & \paramhiggs & \paramletter & \parammnist \\
    \includegraphics[width=0.24\textwidth]{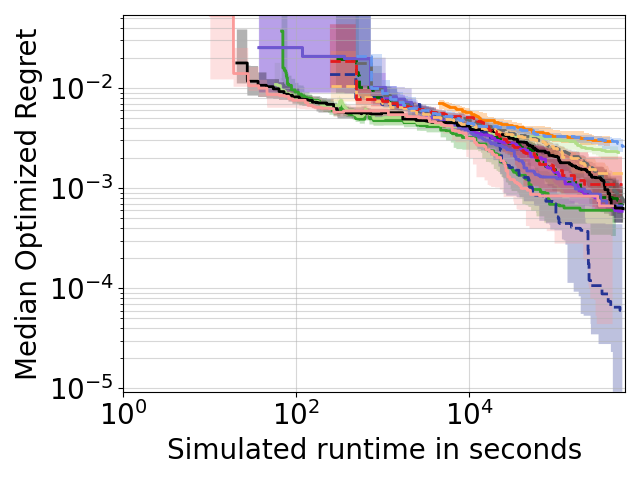} & \includegraphics[width=0.24\textwidth]{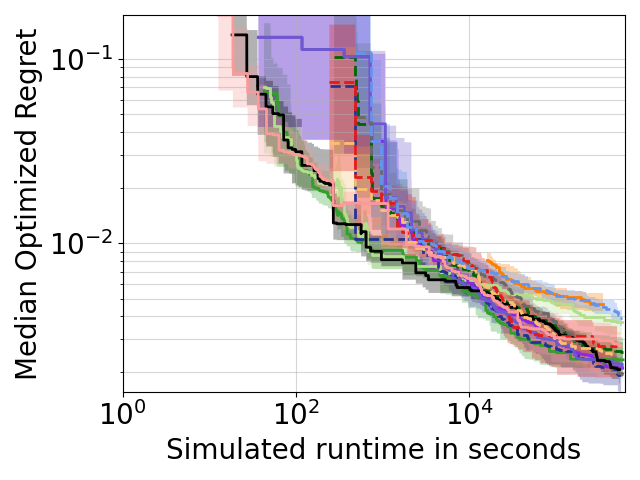} &
    \includegraphics[width=0.24\textwidth]{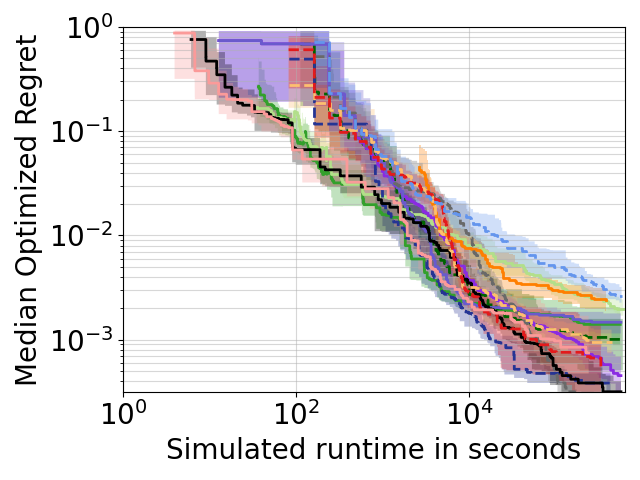} &
    \includegraphics[width=0.24\textwidth]{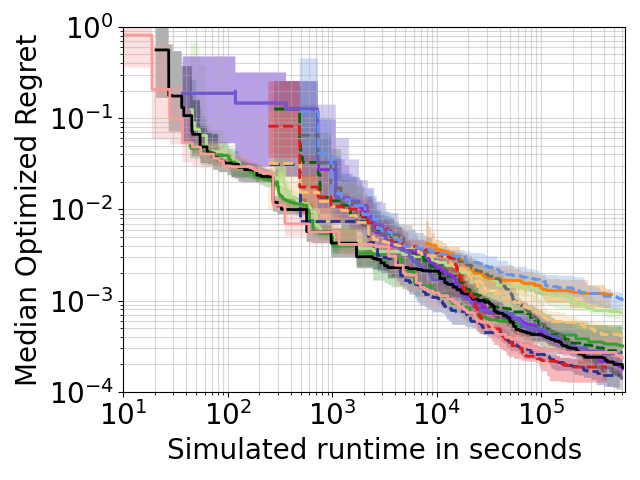} \\
    \paramoptdigits & \parampoker \\
    \includegraphics[width=0.24\textwidth]{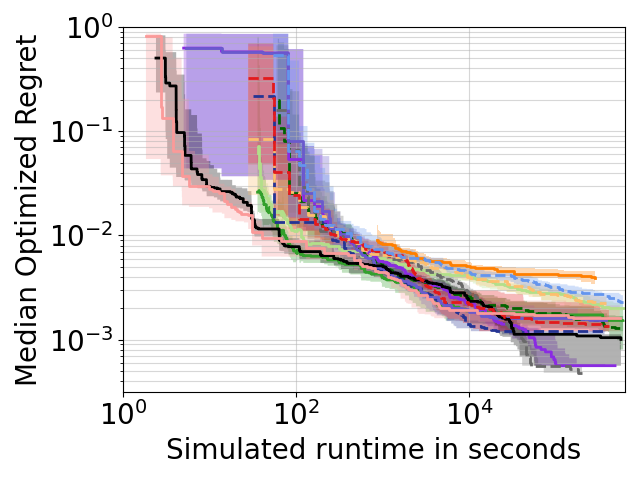} &
    \includegraphics[width=0.24\textwidth]{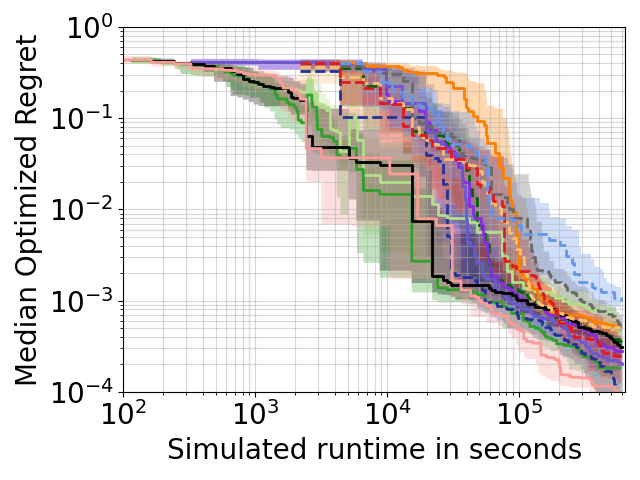} \\
    \NASA & \NASB & \NASC & \\
\end{tabular}
    \caption{Median performance-over-time for \textbf{all} optimizers.
    \label{fig:app:trajall}}
\end{figure}

\begin{figure}[p]
    \centering
    \begin{tabular}{@{\hskip 0mm}c@{\hskip 0mm}c@{\hskip 0mm}c@{\hskip 0mm}c@{\hskip 0mm}}
    \cartpole 
    & \bnnprotein & \bnnyear \\
    \includegraphics[width=0.24\textwidth]{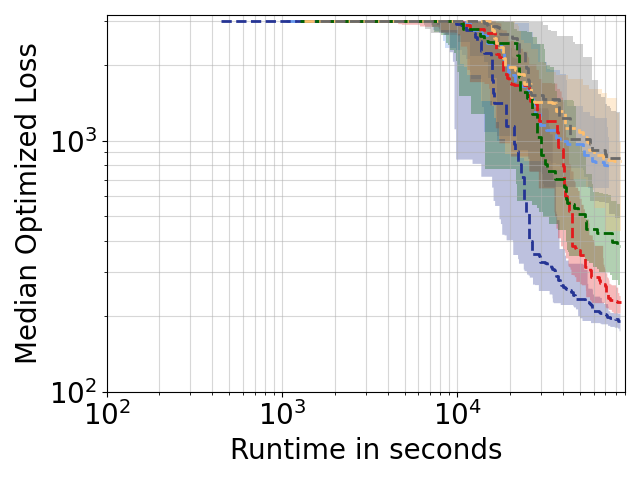} &
    \includegraphics[width=0.24\textwidth]{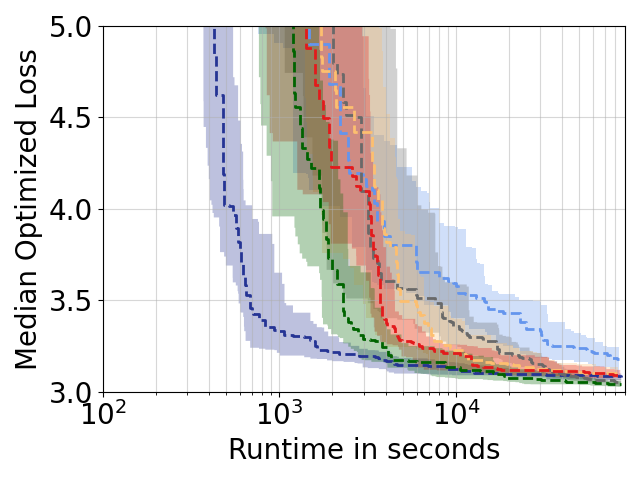} &
    \includegraphics[width=0.24\textwidth]{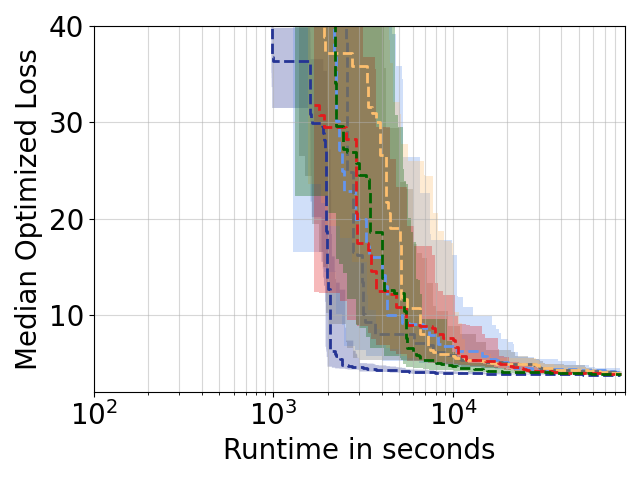} \\
    \paramadult & \paramhiggs & \paramletter & \parammnist \\
    \includegraphics[width=0.24\textwidth]{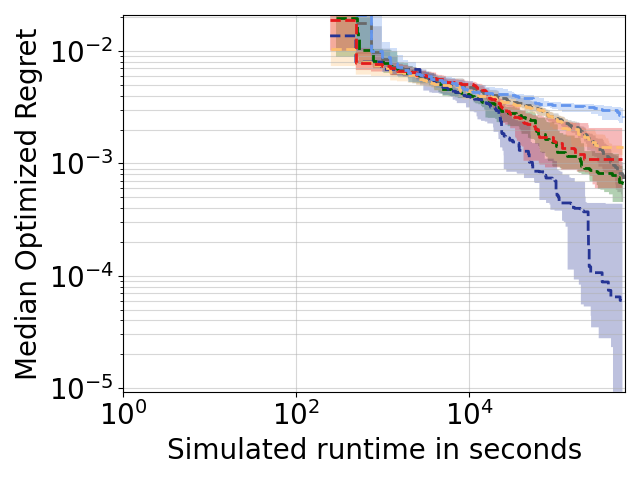} & \includegraphics[width=0.24\textwidth]{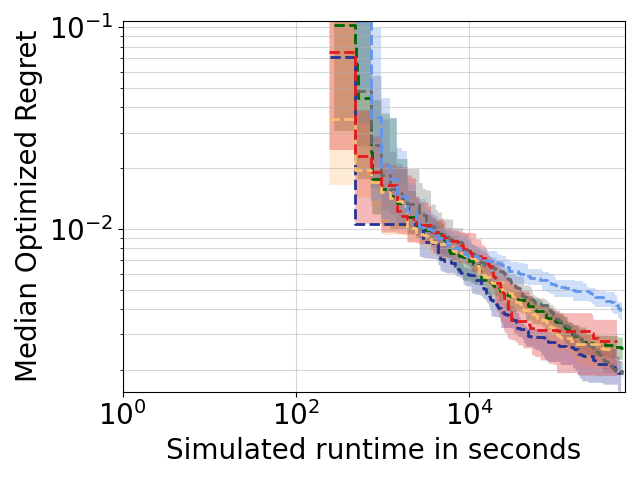} &
    \includegraphics[width=0.24\textwidth]{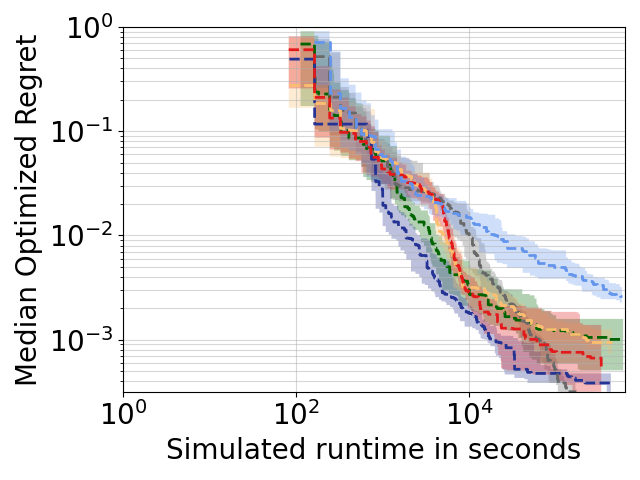} &
    \includegraphics[width=0.24\textwidth]{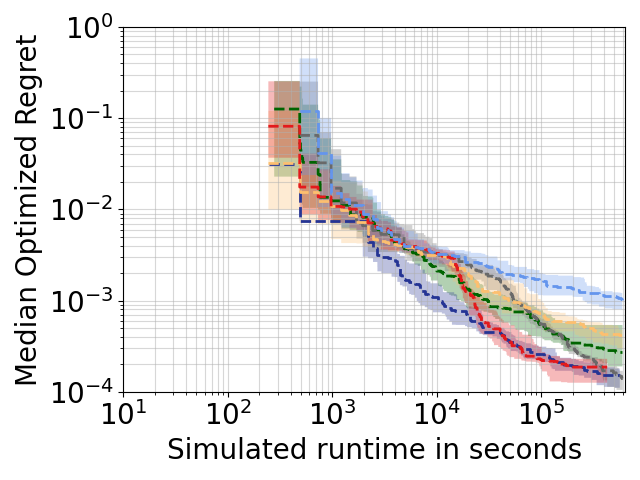} \\
    \paramoptdigits & \parampoker \\
    \includegraphics[width=0.24\textwidth]{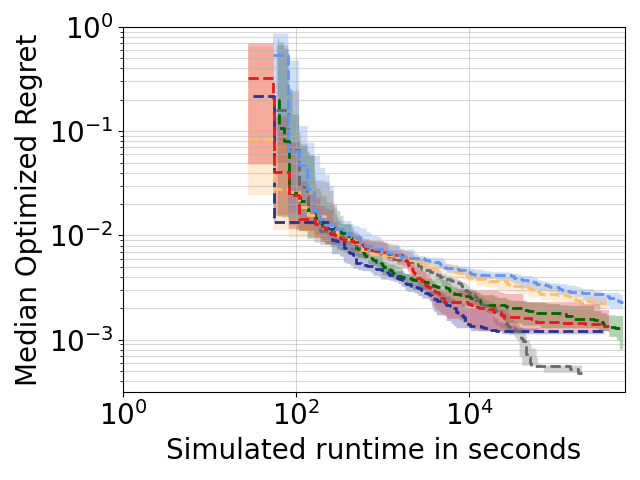} &
    \includegraphics[width=0.24\textwidth]{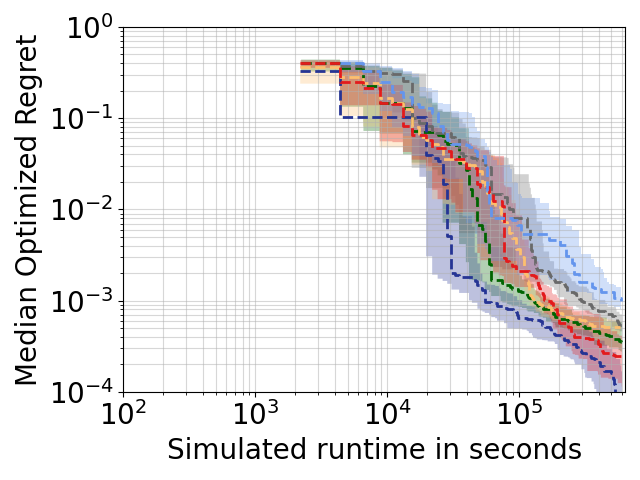} \\
    \NASA & \NASB & \NASC & \\
\end{tabular}
    \caption{Median performance-over-time for \textbf{\bb{}} optimizers.
    \label{fig:app:trajsf}}
\end{figure}

\begin{figure}[p]
    \centering
    \begin{tabular}{@{\hskip 0mm}c@{\hskip 0mm}c@{\hskip 0mm}c@{\hskip 0mm}c@{\hskip 0mm}}
    \cartpole 
    & \bnnprotein & \bnnyear \\
    \includegraphics[width=0.24\textwidth]{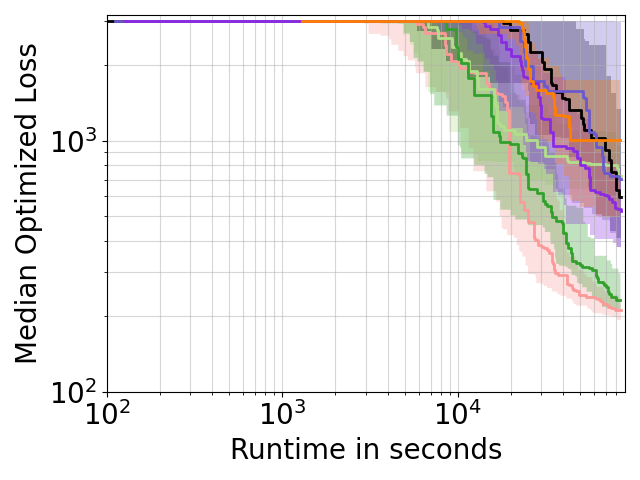} &
    \includegraphics[width=0.24\textwidth]{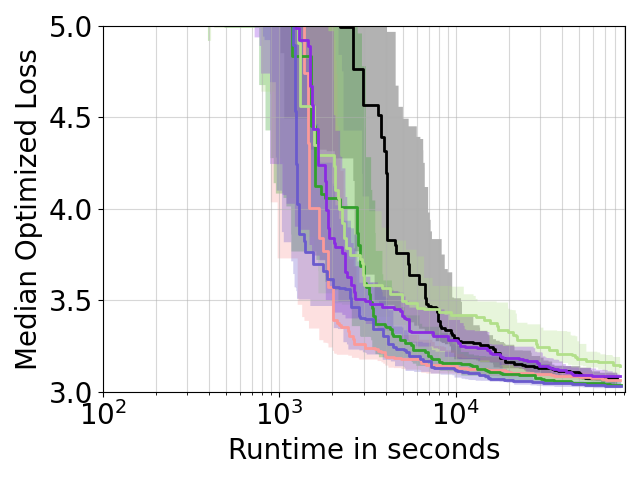} &
    \includegraphics[width=0.24\textwidth]{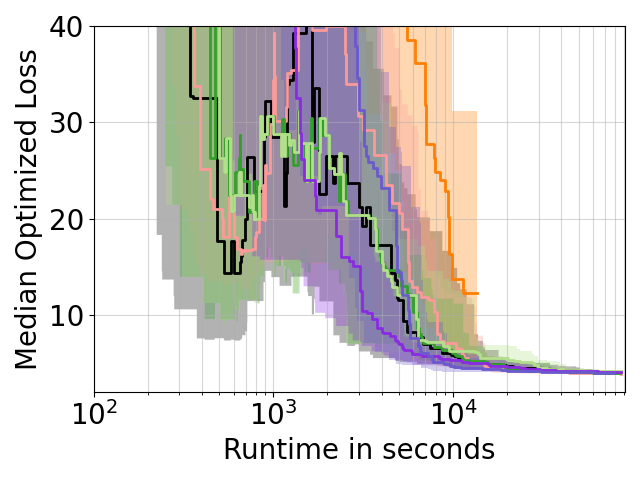} \\
    \paramadult & \paramhiggs & \paramletter & \parammnist \\
    \includegraphics[width=0.24\textwidth]{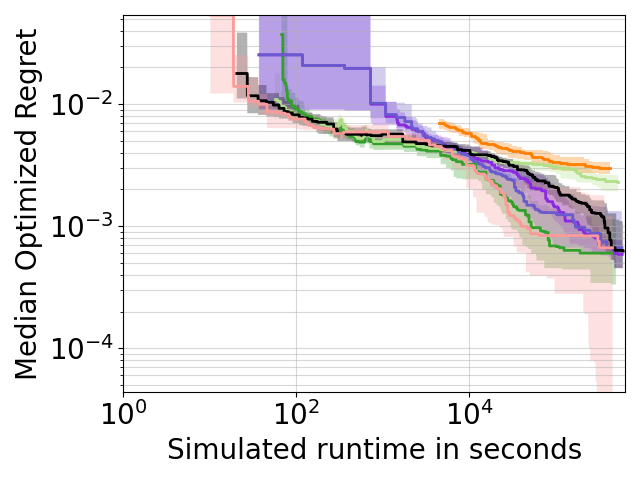} & \includegraphics[width=0.24\textwidth]{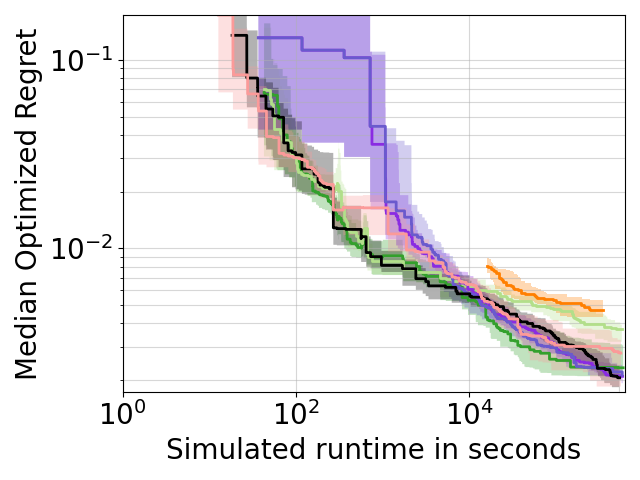} &
    \includegraphics[width=0.24\textwidth]{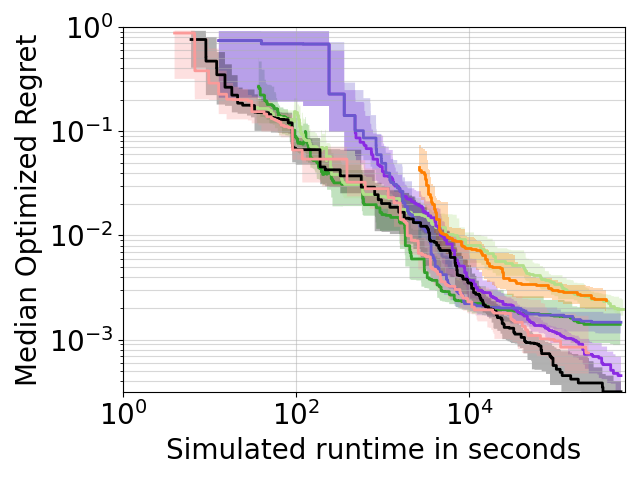} &
    \includegraphics[width=0.24\textwidth]{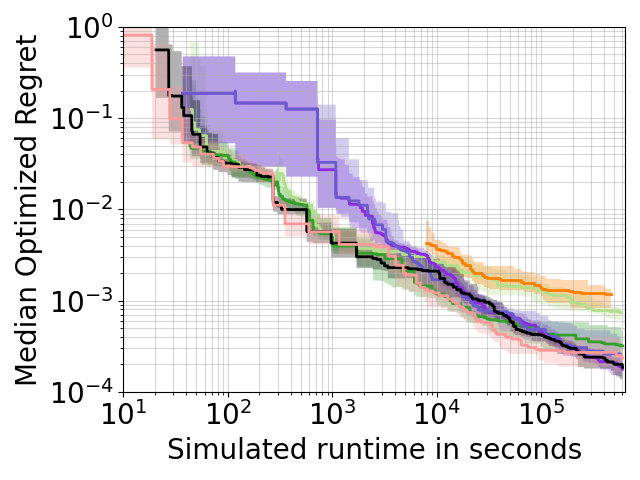} \\
    \paramoptdigits & \parampoker \\
    \includegraphics[width=0.24\textwidth]{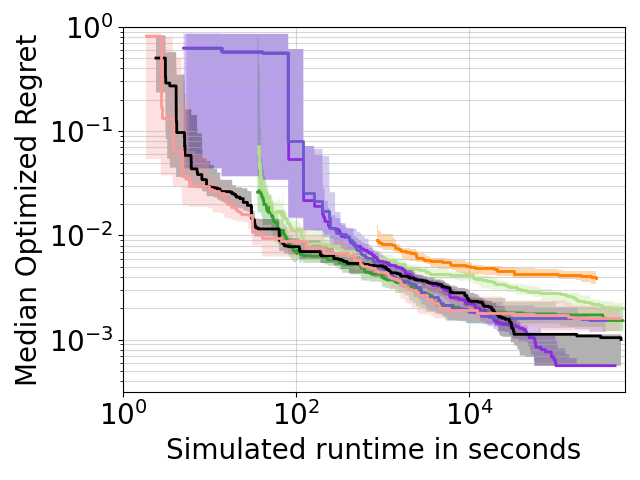} &
    \includegraphics[width=0.24\textwidth]{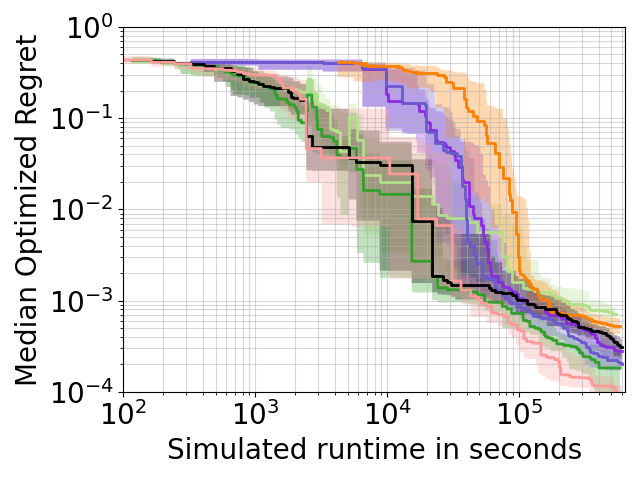} \\
    \NASA & \NASB & \NASC & \\
\end{tabular}
    \caption{Median performance-over-time for \textbf{\mf{}} optimizers.
    \label{fig:app:trajmf}}
\end{figure}

Here, we give more results on our large-scale empirical study. First, we report results for all optimizers in Table~\ref{app:tab:sf},~\ref{app:tab:mf} for the existing community benchmarks, and in Tables~\ref{app:tab-svm-sf}-~\ref{app:tab-nn-mf} for the new benchmarks. Second, we report statistical tests for RQ1 and RQ2 similar to the ones in the main paper for the new benchmarks in Tables~\ref{tab:pvalues-rq1-new} and~\ref{tab:pvalues-rq2-new}. Third, we report average ranking-over-time for each benchmark family in Figure~\ref{fig:app:rankall},~\ref{fig:app:ranksf} and \ref{fig:app:rankmf}. Finally, we show performance-over-time plots for all \emph{existing community} benchmarks in Figure~\ref{fig:app:trajall},~\ref{fig:app:trajsf} and ~\ref{fig:app:trajmf}.

\begin{table}[htbp]
    \centering
    \small
    \caption{\rescaptionexisting{Final performance of each \bb{} optimizer}{}\stattestcaption{}}
    \label{app:tab:sf}
\begin{tabular}{
    @{\hskip 1mm}l@{\hskip 1mm}
    l@{\hskip 2mm}l@{\hskip 2mm}
    l@{\hskip 2mm}l@{\hskip 2mm}
    l@{\hskip 2mm}l@{\hskip 1mm}
    }
    \toprule
benchmark & \multicolumn{6}{l}{\bb{} optimizers} \\
 & \random & \de &  \smacbo & \smacrf & \tpe{} & \hebo \\
\midrule
\cartpole &    786.444 &  851.72222 &  826.444 &  227.056 &      381.72222 &  \textbf{191.833} \\
 \bnnprotein &      3.17763 &  3.05335 &  3.10514 &  3.09424 &  \textbf{3.04537} &  3.08331 \\
 \bnnyear &      4.07933 &  4.01501 &  3.97039 &  3.88006 &        3.86969 &  \textbf{3.77791} \\
 \paramadult &      0.00258 &  0.00072 &  0.00141 &  0.00110 &        0.00068 &  \textbf{0.00006} \\
 \paramhiggs &      0.00390 &  0.00194 &  0.00250 &  0.00277 &        0.00256 &  \textbf{0.00193} \\
 \paramletter &      0.00263 &  \textbf{0.00000} &  0.00095 &  0.00055 &        0.00101 &  0.00038 \\
\end{tabular}

\end{table}

\begin{table}[htbp]
    \centering
    \small
    \caption{\rescaptionexisting{Final performance of each \mf{} optimizer}{}\stattestcaption{}}
    \label{app:tab:mf}
\begin{tabular}{
 @{\hskip 1mm}l@{\hskip 1mm}
 l@{\hskip 2mm}l@{\hskip 2mm}
 l@{\hskip 2mm}l@{\hskip 2mm}
 l@{\hskip 2mm}l@{\hskip 2mm}
 l@{\hskip 1mm}
}
\toprule
benchmark & \multicolumn{7}{l}{\mf{} optimizers} \\
     & \hb{} & \bohb{} & \dehb{} & \smachb{} & \dragonfly & \optunatm{} & \optunath{} \\
 \cartpole &           724.88889 &             232.94444 &  593.83333 &  \textbf{211.33333} &        1004.38889 &         702.33333 &     523.66667 \\
 \bnnprotein &             3.14047 &    3.03529 &  3.07514 &       3.06393 &           9.65112 &  \textbf{3.03252} &       3.08817 \\
 \bnnyear &  4.11971 &    3.92703 &  4.03676 &  \textbf{3.88357} &          12.30007 &  3.91723 &  4.02678 \\
 \paramadult &             0.00232 &    0.00060 &  0.00062 &  0.00067 &           0.00298 &           0.00067 &  \textbf{0.00059} \\
 \paramhiggs &             0.00373 &               0.00232 &  \textbf{0.00206} &       0.00278 &           0.00469 &           0.00212 &  0.00209 \\
 \paramletter &             0.00197 &               0.00140 &  \textbf{0.00032} &       0.00075 &           0.00240 &           0.00147 &       0.00045 \\
\bottomrule
\end{tabular}%
\end{table}

\newpage

\begin{table}[htbp]
    \centering
    \small
    \caption{P-value of a sign test for the hypothesis that advanced methods outperform the baseline \random{} for black-box optimization and \hb{} for \mf{} optimization for the new benchmarks. We underline p-values that are below $\alpha = 0.05$ and also boldface p-values that are below $\alpha = 0.05$ after multiple comparison correction (dividing $\alpha$ by the number of comparisons, i.e. 5 and 4; boldface/underlined implies that the advanced method is better). We also give the wins/ties/losses of \random{} and \hb{} against the challengers.}
    \label{tab:pvalues-rq1-new}
\begin{tabular}{lrrrrr}
        \toprule
        {} & \de & \smacbo & \smacrf & \hebo{} & \tpe{}  \\ 
        p-value against \random{} & $\mathbf{\underline{0.00058}}$ & $\mathbf{\underline{0.00000}}$ & $\mathbf{\underline{0.00000}}$ & $\mathbf{\underline{0.00000}}$ & $0.99909$  \\
        wins/ties/losses against \random{} & $43/33/12$ & $56/27/5$ & $55/24/9$ & $55/22/11$ & $15/30/43$ \\
        \midrule
        {} & \bohb & \dehb & \smachb & \dragonfly & {} \\
        p-value against \hb{} & $0.94535$ & $\mathbf{\underline{0.00000}}$ & $\mathbf{\underline{0.00000}}$ & $0.99992$ \\
        wins/ties/losses against \hb{} & $10/54/24$ & $56/28/4$ & $57/28/3$ & $24/6/58$ & {} \\
        \bottomrule
    \end{tabular}
\end{table}

\begin{table}[htbp]
    \small
    \centering
    \caption{P-values of a sign test for the hypothesis that \mf{} outperform their \bb{} counterparts for the new benchmarks. We boldface p-values that are below $\alpha = 0.05$ (boldface implies that the \mf{} method is better).}
    \label{tab:pvalues-rq2-new}
    \begin{tabular}{lrrrrr}
        \toprule
        Budget & {} &\hb{} vs \random{} & \dehb{} vs \de{} & \bohb{} vs \tpe{} & \smachb{} vs \smacrf{} \\
        \midrule
        \multirow{2}{*}{$100\%$} & p-values & $0.97350$ & $\mathbf{0.02650}$ & $0.99516$ & $0.85820$ \\ 
        & w/t/l & $24/23/41$ & $37/33/18$ & $16/33/39$ & $16/47/25$ \\
        \midrule
        \multirow{2}{*}{$10\%$} & p-values & $\mathbf{0.04284}$ & $\mathbf{0.00000}$ & $0.16870$ & $\mathbf{0.00008}$ \\
         & w/t/l & $50/5/33$ & $68/8/12$ & $43/12/33$ & $57/10/21$ \\
        \midrule
        \multirow{2}{*}{$1\%$} & p-values & $\mathbf{0.00000}$ & $\mathbf{0.00000}$ & $\mathbf{0.00000}$ & $\mathbf{0.00000}$ \\
        & w/t/l & $76/2/10$ & $79/0/9$ & $78/5/5$ & $81/1/6$ \\
        \bottomrule
    \end{tabular}
\end{table}

\newpage
 
\newpage

\end{document}